\PassOptionsToPackage{numbers,sort&compress}{natbib}
\documentclass{article}

\usepackage[preprint]{neurips_2026}

\usepackage[utf8]{inputenc}
\usepackage[T1]{fontenc}
\usepackage{hyperref}
\usepackage{url}
\usepackage{booktabs}
\usepackage{amsfonts}
\usepackage{amssymb}
\usepackage{graphicx}
\usepackage{wrapfig}
\usepackage{makecell}
\usepackage{nicefrac}
\usepackage{microtype}
\usepackage[table,dvipsnames]{xcolor}
\usepackage{amsmath}
\usepackage[nameinlink,capitalise,noabbrev]{cleveref}
\usepackage{multirow}
\usepackage{subcaption}
\usepackage{tcolorbox}
\tcbuselibrary{skins}
\usepackage{wrapfig}
\usepackage{caption}
\usepackage{needspace}
\usepackage{cutwin}
\usepackage{caption}
\usepackage{float}
\usepackage{tabularx}
\newcolumntype{Y}{>{\centering\arraybackslash}X}
\definecolor{ncmetabg}{HTML}{F1F4F7}
\definecolor{ncmetaedge}{HTML}{DCE6F5}
\definecolor{ncmetablue}{HTML}{1877F2}
\definecolor{ncapricot}{HTML}{F7EBDD}
\definecolor{UniMVUHL}{RGB}{232,240,254}
\definecolor{UniMVUAccent}{RGB}{29,78,148}

\newcommand{\NA}{\textemdash}

\newcommand{\best}[1]{\textbf{#1}}
\newcommand{\second}[1]{\underline{#1}}
\newcommand{\oursmethod}[1]{\textbf{\textcolor{UniMVUAccent}{#1}}}

\newcommand{\papercodeurl}{https://tldbn.github.io/UniMVU/}
\newcommand{\paperkeywords}{Large multimodal models, video question answering, multimodal fusion, audio visual reasoning, 3D scene understanding.}

\title{Not All Modalities Are Equal: Instruction-Aware Gating for Multimodal Video Understanding}

\author{%
  \normalfont
  \textbf{Bonan Ding}$^{1}$, \textbf{Umair Nawaz}$^{1}$, \textbf{Ufaq Khan}$^{1}$, \textbf{Abdelrahman M. Shaker}$^{1}$, \\
  \textbf{Muhammad Haris Khan}$^{1}$, \textbf{Jiale Cao}$^{2}$, \textbf{Jin Xie}$^{3}$, \textbf{Fahad Shahbaz Khan}$^{1,4}$ \\[2pt]
  $^{1}$Mohamed bin Zayed University of Artificial Intelligence \quad
  $^{2}$Tianjin University \\
  $^{3}$Chongqing University \quad
  $^{4}$Linköping University \\[2pt]
  \texttt{bonan.ding@mbzuai.ac.ae}
}

\makeatletter
\newcommand{\maketitleboxed}[1]{%
  \par
  \begingroup
    \renewcommand{\thefootnote}{\fnsymbol{footnote}}
    \renewcommand{\@makefnmark}{\hbox to \z@{$^{\@thefnmark}$\hss}}
    \long\def\@makefntext##1{%
      \parindent 1em\noindent
      \hbox to 1.8em{\hss $\m@th ^{\@thefnmark}$}##1%
    }
    \thispagestyle{empty}%
    \begin{tcolorbox}[
      enhanced,
      colback=ncmetabg,
      colframe=ncmetaedge,
      boxrule=0.35pt,
      arc=12pt,
      left=0.55cm, right=0.55cm, top=0.45cm, bottom=0.4cm,
      interior style={shade, shading angle=315,
        left color=white!96!ncmetabg,
        right color=ncmetablue!4!ncapricot!8!ncmetabg},
      before skip=0pt, after skip=0.4em,
      grow to left by=1.5pt, grow to right by=1.5pt,
    ]
      \centering
      {\LARGE\bf \@title\par}%
      \vskip 0.22in
      \def\And{\end{tabular}\hfil\linebreak[0]\hfil\begin{tabular}[t]{c}\bf\rule{\z@}{24\p@}\ignorespaces}%
      \def\AND{\end{tabular}\hfil\linebreak[4]\hfil\begin{tabular}[t]{c}\bf\rule{\z@}{24\p@}\ignorespaces}%
      \begin{tabular}[t]{c}\bf\rule{\z@}{24\p@}\@author\end{tabular}\par
      \vskip 0.18in
      \begingroup
        \leftskip=1.5em \rightskip=1.5em
        \centerline{\large\bf Abstract}\vspace{0.6ex}
        \small #1\par
        \vspace{0.6ex}
        {\parfillskip=0pt plus 1fil\relax
         \noindent\small\textbf{Project page:}~\href{\papercodeurl}{\color{blue}\texttt{\papercodeurl}}\par
         \noindent\small\textbf{Keywords:}~\paperkeywords\par}
      \endgroup
    \end{tcolorbox}%
    \@thanks
    \@notice
  \endgroup
  \let\maketitle\relax
  \let\thanks\relax
}
\makeatother

\begin{document}

\maketitleboxed{Pre-trained video large language models excel at visual reasoning. However, they struggle when videos arrive with auxiliary streams, such as audio, depth map, or dense temporal evidence. In such a scenario, uniform fusion induces \emph{modality interference}, allowing irrelevant channels to distract the model. To address this issue, we present a unified multimodal video understanding framework, named UniMVU, that performs instruction-aware fusion across video, audio, depth map, or any other modality inputs via two levels of dynamic gating: inner-modality gates emphasize salient regions within each modality, whereas modality-level gates re-weight whole streams; both are conditioned on the text instruction to adaptively balance modality importance. Our UniMVU combines cross-modal self-attention with instruction-driven inner-modality gating module and a modality-level gating module with control token; for time-aligned streams we further adopt a fast-to-slow fusion scheme that reduces redundancy. Across six benchmarks (AVQA, AVSD, Music-AVQA, ScanQA, SQA3D and MVBench), our UniMVU achieves consistent gains over static-fusion baselines achieving gains as high as 13.5 in terms of CIDEr metric. Further, our analysis shows that the gating mechanism aligns with the human-interpretable modality relevance, and ablations show the contributions of inner-modality and modality-level gating. Our UniMVU provides a simple, unified recipe for instruction-aware multimodal video understanding that scales to diverse modalities without hand-crafted fusion rules.
}


\begin{wrapfigure}{r}{0.6\linewidth}
    \centering
    \vspace{-8pt}
    \includegraphics[width=\linewidth]{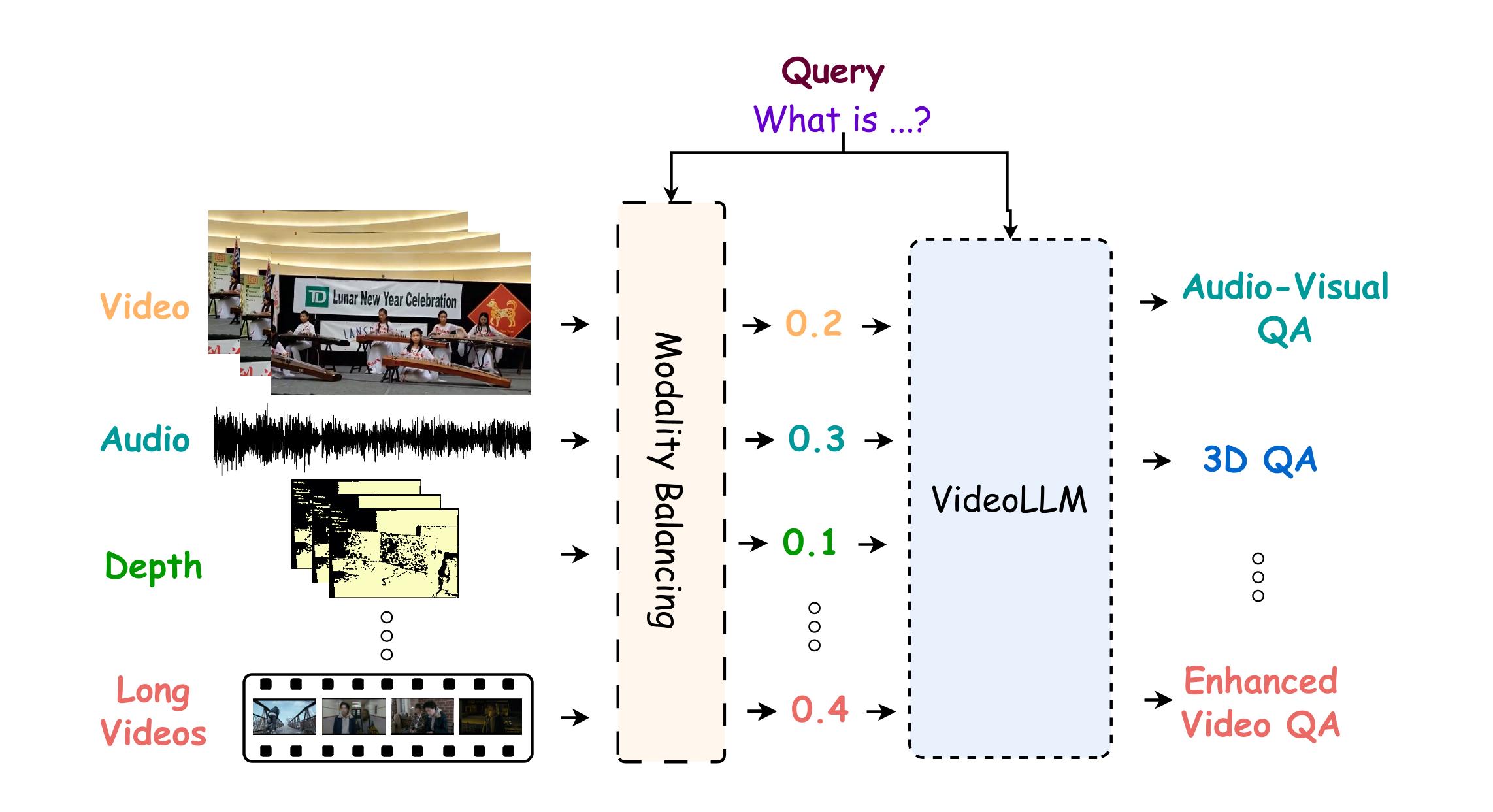}
    \caption{Context-dependent modality weighting in UniMVU. For each sample, UniMVU assigns query-conditioned weights to video, audio, depth, or long-video streams so that the Video LLM emphasizes the evidence required by the instruction}
    \label{fig:teaser-1}
    \vspace{-8pt}
\end{wrapfigure}

\section{Introduction}
\label{sec:intro}

Recent advances in Large Language Models (LLMs) have achieved strong generalization in natural language understanding and generation, motivating a natural progression toward video large language models (Video LLMs) that connect pretrained visual or video encoders with instruction-following LLMs through projection layers, query-formers, or cross-modal interfaces~\cite{gpt3,flant5,llama,qwen2,llava_onevision,VideoChatGPT,videollama}. By embedding sampled frames as language-compatible tokens, Video LLMs provide an interactive interface for open-ended video understanding: given a video and a natural-language instruction, they are expected to recognize objects, actions, and events; answer questions; describe temporal dynamics; and reason about spatial layout, causal relations, and commonsense situations in the video. Compared with task-specific video models, this LLM-centered paradigm supports a broader range of flexible tasks, including video question answering, video dialogue, event captioning, temporal grounding, and instruction-following video reasoning. This RGB--text recipe is strong for many appearance-centric questions, but it leaves an important gap: the evidence needed to answer a user instruction is often distributed across synchronized modalities. Sound can identify speakers, instruments, and off-screen events; depth or point-cloud cues can resolve spatial layout; and denser temporal streams can capture actions whose decisive evidence appears between sparsely sampled frames~\cite{videollama2,cat,cdViews,scesu}. Simply appending these streams as extra tokens leaves their relative importance to be discovered implicitly during decoding.

That importance is query-dependent rather than fixed by the modality or dataset. Audio is essential for questions about speech, loudness, or instruments, but it can become nuisance evidence for recognizing color or counting visible objects. Depth and 3D cues help resolve indoor layout and spatial relations, yet they may add little to appearance-centric questions; dense temporal features capture brief actions that sparse frames miss, but can be redundant for short clips or static object queries. Fixed or uniform fusion therefore risks \emph{modality interference}, in which auxiliary evidence competes with or distracts from the cues required by the current instruction~\cite{videollama2,cat,pave}. This observation motivates a fusion stage that is conditioned on the instruction before answer generation, so that the model can select informative tokens within each stream and assign appropriate global weight to each stream.

\begin{figure*}[t]
    \centering
    \includegraphics[width=\textwidth]{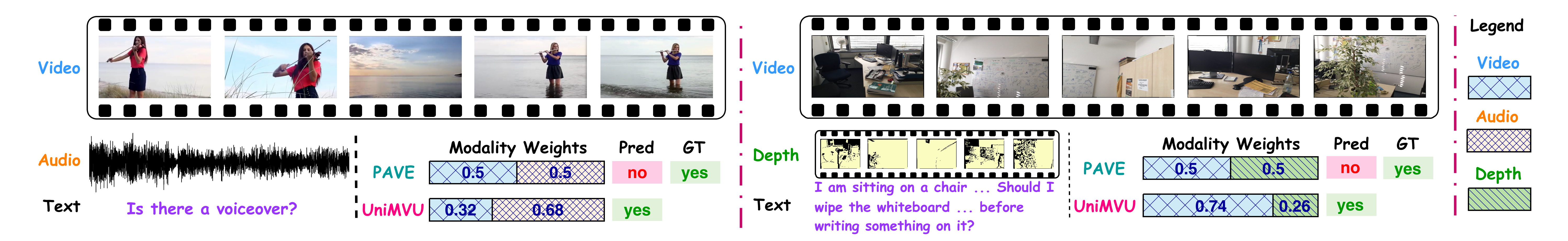}
    \caption{Qualitative comparison between UniMVU and uniform weighting. UniMVU adapts modality weights to the input query, while PAVE~\cite{pave} uses uniform weighting. In the audio-centric example, UniMVU upweights audio and matches the ground truth; in the 3D-VQA example, UniMVU emphasizes visual evidence over depth and answers correctly. Additional examples are shown in Figs.~\ref{fig:qualitative}--\ref{fig:qualitative_additional_2}.}
    \label{fig:teaser-2}
\end{figure*}

Existing Video LLM extensions only partially address this issue. Many methods incorporate auxiliary streams through fixed projectors, cross-attention layers, side-channel tokens, or task-specific interaction modules~\cite{flamingo,videollama,videollama2,pave}. Query-aware and delegation-style schemes further improve fusion for particular settings~\cite{pave,cat,videollama2,avllm}, but they often leave stream-level relevance to be handled implicitly by decoder attention or design the interaction around a specific modality pair. Consequently, the model still lacks an explicit answer to two instruction-dependent questions: which tokens within a stream should be emphasized, and which stream should contribute more to the final representation.

This gap becomes more evident when moving beyond a single auxiliary-modality benchmark. The additional evidence may be audio, depth/3D, or dense temporal video tokens, and these streams differ in temporal granularity, reliability, and alignment with the question. A unified model therefore cannot rely on a separate connector for audio--visual QA, another for RGB-D scene QA, and another for dense VideoQA. It needs a common fusion rule that can transfer across modality combinations while still respecting the specialization of frozen encoders.

To address these gaps, we propose UniMVU (see Fig.~\ref{fig:teaser-1}), a unified multimodal video understanding framework for gating video, audio, depth/3D, and dense-video streams before LLM decoding. UniMVU uses cross-modal self-attention to obtain instruction-conditioned interactions and then applies two residual gates. The inner-modality gate computes token-level relevance within each stream, while the modality-level gate compares complete streams through a shared learnable control token. Unlike hard gating, the gated signal is added as a refinement to the attended representation; this keeps the original evidence available while increasing the relative influence of instruction-relevant tokens and streams.

\subsection{Contributions}
\begin{wrapfigure}{r}{0.5\linewidth}
    \centering
    \vspace{-8pt}
    \includegraphics[width=\linewidth]{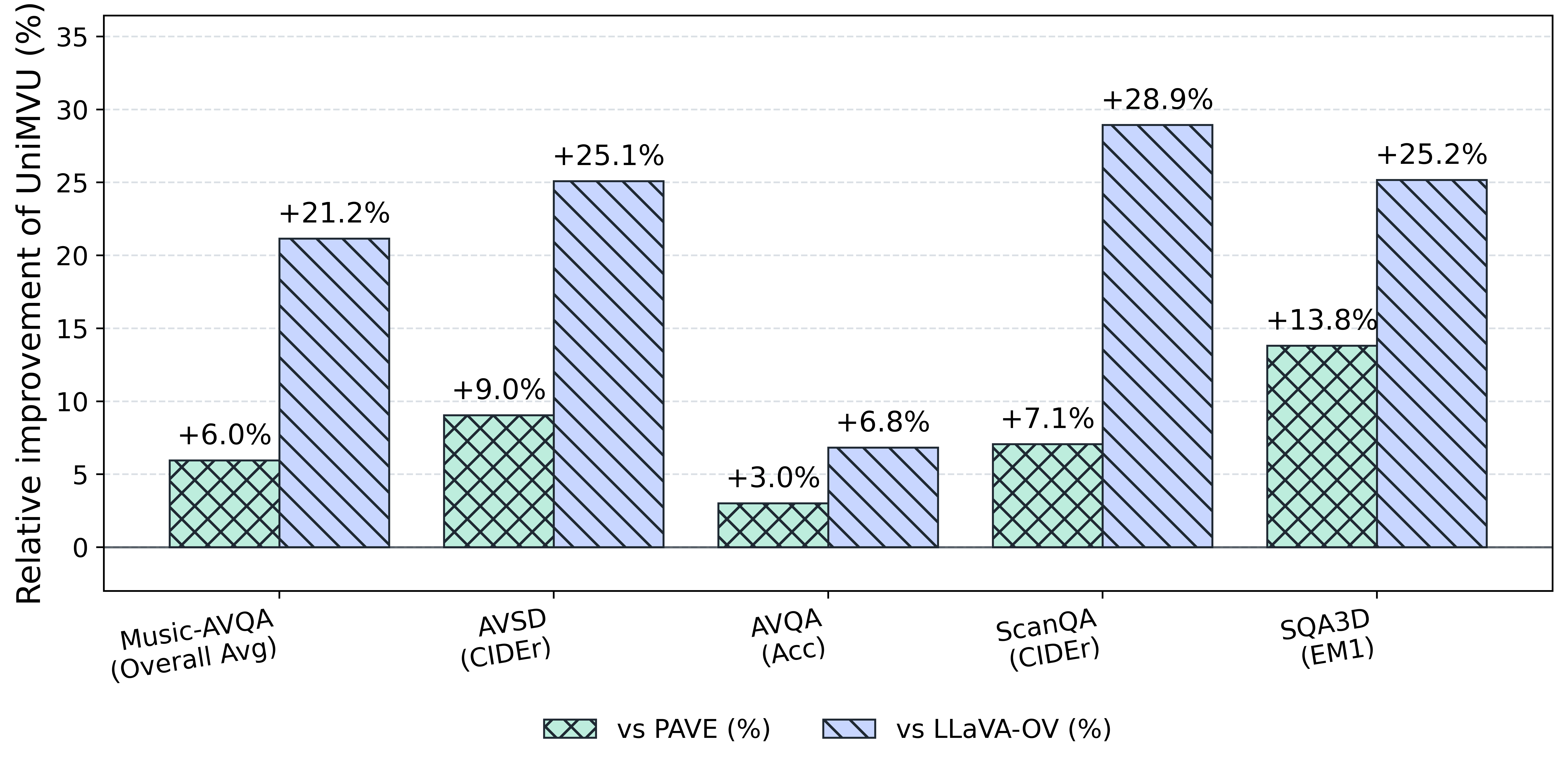}
    \caption{Relative improvements of UniMVU over PAVE~\cite{pave} and LLaVA-OV~\cite{llava_onevision} across representative benchmarks. The plot highlights aggregate trends; detailed tables report the metric-specific gains and the subtask trade-offs observed under unified training.}
    \label{fig:teaser-3}
    \vspace{-8pt}
\end{wrapfigure}

We present \textbf{UniMVU}, a unified large multimodal video understanding model that ingests mixed-modality inputs including video, audio, and depth/3D or any other modality together with text instruction, and enables \emph{jointly} training across modality-mixed datasets and tasks, enabling a single model to generalize across diverse modality configurations.
The core of UniMVU is an instruction-driven question-conditioned dynamic gating mechanism that operates at two levels i.e., feature and modality, allowing the model to tailor fusion to each query and to overcome the limitations of fixed or single-level fusion. The learned gates explicitly re-balance modalities on a per-query basis (e.g., up-weighting audio for “what was said?” and combining RGB with depth for “what action is occurring?”), and we show that their behavior aligns with human-interpretable notions of modality relevance. We validate UniMVU on a broad suite of audio-visual and 3D-enhanced video-QA benchmarks, including AVQA, AVSD, Music-AVQA, ScanQA, SQA3D and MVBench, demonstrating consistent improvements over traditional fusion baselines and strong results on modality-specific queries (see Fig.~\ref{fig:teaser-3}).


\section{Related Work}
\label{sec:relate}

\subsection{Video and Multimodal LLMs}
Large language models have achieved strong generalization in natural language understanding and generation, motivating a rapidly growing family of multimodal large language models that connect pretrained perceptual encoders with LLMs~\cite{gpt3,flant5,llama,qwen2}. Flamingo~\cite{flamingo} introduces cross-attention layers to bridge pretrained vision and language models for few-shot multimodal learning. Video-ChatGPT~\cite{VideoChatGPT} extends instruction tuning to detailed video understanding. Video-LLaMA~\cite{videollama} incorporates visual and audio encoders for video-language interaction, while VideoLLaMA~2~\cite{videollama2} improves spatial-temporal modeling and audio understanding in Video LLMs. LLaVA-OneVision~\cite{llava_onevision} studies visual task transfer across images and videos, and MovieChat+~\cite{moviechat_plus_tpami} introduces question-aware sparse memory for long video question answering. Recent TPAMI studies further extend multimodal LLMs through in-context instruction tuning and mixture-of-experts scaling~\cite{otter_tpami2025,unimoe_tpami2025}. These works establish LLM-centered architectures as a powerful interface for open-ended visual and video reasoning.

Despite this progress, most Video LLMs first construct a fixed set of input tokens and then rely on the decoder to use the relevant evidence implicitly. This design is effective when the input is dominated by sampled RGB frames, but becomes less direct when videos are accompanied by heterogeneous auxiliary streams. Audio, depth/3D features, and dense temporal tokens can all provide useful cues, while their relevance changes substantially across questions. Therefore, multimodal video understanding requires not only attaching additional encoders to an LLM, but also balancing the contribution of different evidence sources according to the current instruction.

\subsection{Video Question Answering}
Video question answering has been widely studied as the problem of connecting language queries with spatial and temporal video evidence~\cite{xue2017unifying,webvideo_vqa_tpami,moviechat_plus_tpami,paa_videoqa_tpami,contrastive_videoqa_tpami2023,invariant_grounding_videoqa_tpami2025}. Early open-ended VideoQA methods learn joint video-question attention to associate questions with spatio-temporal representations~\cite{xue2017unifying}. 
Learning to Answer Visual Questions from Web Videos~\cite{webvideo_vqa_tpami} studies scalable generation of video-question-answer pairs from narrated web videos.
MovieChat+~\cite{moviechat_plus_tpami} uses question-aware memory consolidation to preserve long-range evidence for long video QA. Parse, Align and Aggregate~\cite{paa_videoqa_tpami} parses questions, aligns them with visual evidence, and aggregates graph-based compositional reasoning.
Other VideoQA studies improve grounding through video graph contrastive learning~\cite{contrastive_videoqa_tpami2023} and transformer-based invariant grounding~\cite{invariant_grounding_videoqa_tpami2025}.

Related VQA studies also highlight the importance of relation modeling, causal reasoning, and robustness. MRA-Net~\cite{mranet_tpami} models multimodal relations for visual question answering, cross-modal causal reasoning methods study event-level VQA~\cite{crossmodal_causal_vqa_tpami2023}, and robust VQA studies~\cite{robustvqa_tpami} show that models may exploit language priors or dataset biases when answers are weakly grounded. These methods demonstrate the value of question-conditioned evidence selection, but their selection mechanisms mainly operate over visual tokens, frames, memories, or graph nodes. They provide limited treatment of the case where separate modality streams must be compared and fused under the same language instruction.

\subsection{Multimodal Video Representation and Fusion}
Learning representations that connect video with additional modalities has received considerable attention~\cite{imagebind,languagebind,vast,pave}. ImageBind~\cite{imagebind} learns a shared embedding space across image, text, audio, depth, thermal, and IMU signals. LanguageBind~\cite{languagebind} extends video-language pretraining to multiple modalities through language-centered alignment. VAST~\cite{vast} studies vision-audio-subtitle-text pretraining for omni-modality understanding. PAVE~\cite{pave} adapts pretrained Video LLMs to side-channel signals such as audio, 3D cues, multi-view videos, and high-frame-rate videos. These works make heterogeneous streams easier to connect to language models, but alignment or token insertion alone does not determine how much each stream should contribute to a particular question.

Different downstream tasks further demonstrate the need for auxiliary modality information. Audio-visual QA uses sound and visual events~\cite{avqa,music-avqa,pstpnet,avafnet,avmaster,cat,catplus_tpami2025,avllm,videosalmonn,videosalmonn2}. 3D question answering uses depth, point clouds, scene representations, or selected views~\cite{scanqa,sqa3d,3dllm,pointllm,scenellm,llava3d,threeur_llm,cdViews,jm3d_tpami2025,gen3dvlm_tpami2025}. Dense and long-video benchmarks stress temporal evidence beyond sparse frames~\cite{mvbench,videomme,longvideobench,mlvu}. CAT/CAT+~\cite{cat,catplus_tpami2025} improves question-related audio-visual interaction, while modality-collaborative models such as mPLUG-Owl2~\cite{owl2} study interaction among modalities. Recent analyses further suggest that irrelevant or weakly aligned streams can cause modality interference in multimodal LLMs~\cite{diagnosing,llava_av_ssm}. 
Overall, prior work has advanced modality alignment, task-specific fusion, and benchmark construction, but explicit instruction-conditioned stream reweighting remains underexplored across multimodal video inputs.


\section{Method}
\label{sec:method}

\begin{figure*}[t]
    \centering
    \includegraphics[width=\textwidth]{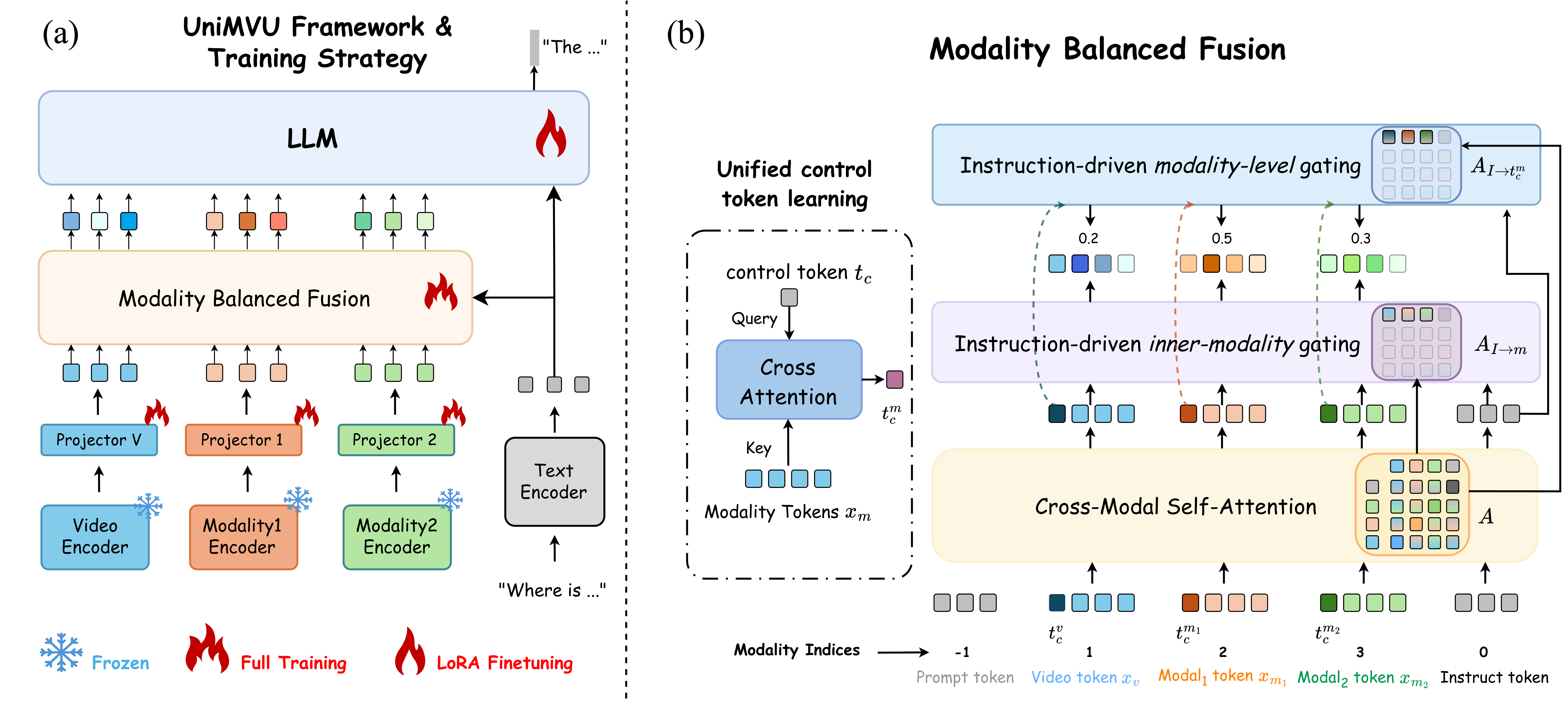}
    \caption{Overview of UniMVU. (a) UniMVU uses modality-specific encoders and projectors to extract features from the supported streams available in a sample, including RGB/video, audio, depth/3D, and dense-video features. The text instruction and modality tokens are processed by the Modality Balanced Fusion block, and the fused tokens are passed to a pretrained LLM for answer generation. During training, modality encoders are frozen, projectors and the fusion block are fully trained, and the LLM is adapted with LoRA. (b) The Modality Balanced Fusion block consists of cross-modal self-attention, instruction-driven inner-modality gating, and instruction-driven modality-level gating.}
    \label{fig:overview}
\end{figure*}

Our approach unifies multimodal video understanding by coupling modality-specific encoders and projectors with an instruction-aware fusion module that precedes a pretrained LLM. The fusion module contains three components:
(1) \textbf{Self-Attention for Cross-Modal Interaction} enables token-level interactions among modalities and instruction tokens; 
(2) \textbf{Instruction-Driven Inner-Modality Gating} summarizes each modality into an instruction-selected token set by reweighting tokens according to instruction-guided relevance; and
(3) \textbf{Instruction-Driven Modality-Level Gating} assigns a attention-derived, instruction-conditioned factor to each modality via a dedicated control token, thereby modulating the contribution of each modality. The fused tokens are then passed to the LLM for generating answers.

\subsection{Self-Attention for Cross-Modal Interaction}
\label{subsec:method1}
\textbf{Unified Token Sequence Construction.}
We first project all available inputs into a common embedding space and merge them into a single token sequence alongside the text instruction tokens and system prompts. The available streams depend on the dataset: audio--visual QA uses RGB/video and audio tokens, 3D QA uses RGB/video together with depth or 3D scene tokens, and VideoQA uses sampled RGB/video tokens together with dense temporal tokens. For each non-text modality $m$, given encoder features $\mathbf{F}_m \in \mathbb{R}^{T_m\times d_m}$, where $T_m$ is the number of tokens produced by the corresponding frozen encoder, we apply a learned projector $P_m: \mathbb{R}^{d_m} \to \mathbb{R}^{d}$ to map them to the hidden dimension: $\mathbf{X}_m=P_m(\mathbf{F}_m)\in\mathbb{R}^{T_m\times d}$.  
We flatten each modality features into a token sequence and concatenate it with the instruction embeddings $\mathbf{X}_{\text{text}}$ from the LLM tokenizer:
\begin{equation}
\mathbf{H} = [\mathbf{X}_{\text{text}}; \mathbf{t}_c^1; \mathbf{X}_1; \cdots; \mathbf{t}_c^{N_m}; \mathbf{X}_{N_m}],
\end{equation}
where $N_m$ denotes the number of available non-text modalities for the current sample and $\mathbf{t}_c^m$ is the control token used for modality-level gating. 
Missing modalities are represented by masked placeholder zero tokens and are excluded from attention and gate normalization.
System prompt and padding tokens are used only to form mini-batches and are masked in attention and gate computation.

To distinguish modality identity after projection, we maintain a modality index for every token: $m_i = 0$ for instruction tokens, $m_i = r > 0$ for tokens from the $r$-th available non-text modality, and $m_i = -1$ for padding tokens and system prompts. This unified sequence representation lets subsequent attention layers reason jointly over text and all available streams without introducing separate fusion heads for each task family. It also prevents modality identity from being lost after projection: although all tokens live in the same hidden space, the index preserves which tokens should be normalized together during inner-modality gating and which tokens should contribute to a stream-level score.

\textbf{Cross-Modal Self-Attention Layer.}
To enable rich cross-modal interactions and prepare for the following instruction-aware weighting, we introduce a self-attention-based cross-modal interaction layer operating on the unified sequence $\mathbf{H}$. This layer is built on multi-head self-attention~\cite{attention,qwen2}. Let $\mathbf{H} \in \mathbb{R}^{L\times d}$ denote the input unified sequence with $L$ tokens. We first compute the projections of query, key, and value for each token as:
\begin{equation}
\begin{aligned}
\mathbf{Q} &= \mathbf{H} W_Q,\quad
\mathbf{K} = \mathbf{H} W_K,\quad
\mathbf{V} = \mathbf{H} W_V,
\end{aligned}
\end{equation}
where $W_Q, W_K, W_V \in \mathbb{R}^{d\times d}$ are learned projection parameters. We then apply rotary positional embeddings~\cite{rope} to the projected queries and keys to encode their positions. Multi-head attention is computed independently for each head as:
\begin{equation}
\mathbf{A}^{(h)} = \mathrm{softmax}\Big( \frac{\mathbf{Q}^{(h)}{\mathbf{K}^{(h)}}^\top}{\sqrt{d_h}} + \mathbf{M}_{\text{attn}} \Big),
\label{eq:attn}
\end{equation}
where $d_h$ is the dimension of each attention head and $\mathbf{M}_{\text{attn}}$ masks padding and system-prompt positions that should not participate in gating. Instruction tokens remain visible to non-text modality tokens because they provide the query context for gating. 
Rows of $\mathbf{A}\in\mathbb{R}^{n_h\times L\times L}$ correspond to query positions and columns correspond to key positions; modality-token queries condition the representations on instruction tokens, whereas instruction-token queries attending to modality or control-token keys provide the scores used for gating.
The head outputs are concatenated and projected to obtain:
\begin{equation}
\label{eq:selfout}
\mathbf{O} = \mathrm{Concat}_{h=1}^{n_h}(\mathbf{A}^{(h)}\mathbf{V}^{(h)})W_O,
\end{equation}
where $W_O\in\mathbb{R}^{d\times d}$ is the output projection.
Therefore, the output $\mathbf{O} \in \mathbb{R}^{L \times d}$ contains cross-modal interactions among all visible tokens.. Afterward, we perform Instruction-Driven Inner-Modality Gating (see \S~\ref{subsec:method2}) and Instruction-Driven Modality-Level Gating (see \S~\ref{subsec:method3}), which can adaptively fuse the information from different modalities. Then, we add a residual connection and layer normalization to the attention output, followed by a feed-forward network (FFN) and another residual addition. This produces final updated token representations $\hat{\mathbf{H}}$. To preserve the integrity of the instruction tokens, we propose to keep instruction embeddings untouched. In particular, after the attention and FFN layers, we replace the output for any system prompt and instruction token with its original input embedding. Formally, if $\mathbf{H}$ is the input to this cross-modal fusion module,  the output $\hat{\mathbf{H}}_{\text{out}}$ for each position $i$ can be calculated as
\begin{equation}
\hat{\mathbf{H}}_{\text{out},i} =
\begin{cases}
\mathbf{H}_{i}, & \text{if } m_i \in \{0, -1\},\\
\hat{\mathbf{H}}_i, & \text{otherwise},
\end{cases}
\end{equation}
where $m_i$ is the modality index of token $i$. In words, modality tokens absorb information from the instruction through $\hat{\mathbf{H}}$, while instruction and padding tokens are restored to their input embeddings. This ensures a one-way flow of information: modality tokens are conditioned on the textual instruction, while the instruction representation itself remains unchanged and insulated from modality-specific noise. This separation is critical for stabilizing LLM training, as it prevents noisy non-textual modality features from directly perturbing the instruction embeddings of the well-pretrained LLM.

\begin{figure*}[!t]
    \centering
    \includegraphics[width=\textwidth]{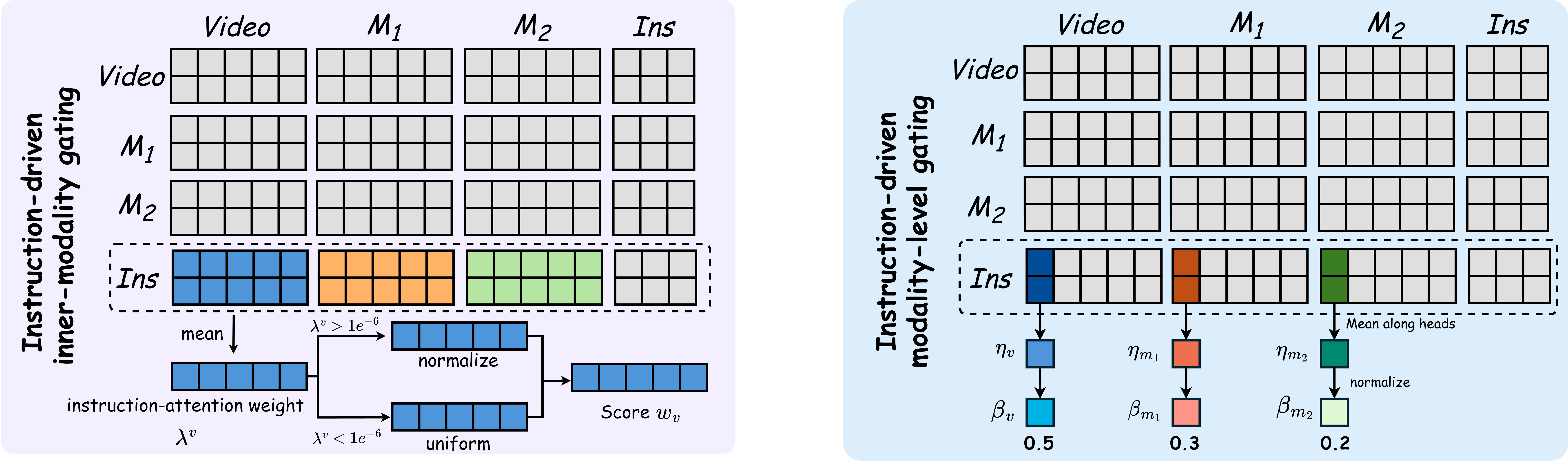}
    \caption{Instruction-driven inner-modality and modality-level gating. The grids denote token-level attention matrices partitioned by modality, where rows are query tokens and columns are key tokens. Left: inner-modality gating uses instruction-to-modality attention to compute normalized token weights within each modality. Right: modality-level gating uses instruction-to-control-token attention, where the control token is the first token in each modality sequence, to produce one scalar weight for each modality.}
    \label{fig:modality-gating}
\end{figure*}

\subsection{Instruction-Driven Inner-Modality Gating}
\label{subsec:method2}

To better fuse the features within each modality that are relevant to the instruction, we introduce an inner-modality gating mechanism to refine each modality's token sequence using instruction-guided weighting as shown in Fig.~\ref{fig:modality-gating} left.
Specifically, we leverage the attention tensor $\mathbf{A}$ from Eq.~\ref{eq:attn} to estimate the importance of each token with respect to the instruction. Let $\mathcal{I}$ be the set of instruction tokens and $\mathcal{M}_m$ be the set of content tokens from modality $m$, excluding the control token $\mathbf{t}_c^m$, which is used for stream-level scoring. 
For the $t$-th token $x_t^m \in \mathcal{M}_m$ in modality $m$, we quantify its instruction relevance using the cross-modal self-attention map. Specifically, we aggregate the attention mass from all instruction tokens to this token over all $n_h$ attention heads, and define the resulting aggregated instruction-attention weight as:
\begin{equation}
\lambda_t^m = \frac{1}{n_h} \sum_{h=1}^{n_h} \sum_{i \in \mathcal{I}} \mathbf{A}_{i \rightarrow t}^{(h)}.
\end{equation}
We then normalize the token weights within modality $m$ as:
\begin{equation}
\label{eq:inner-gating}
w_{m,t} =
\begin{cases}
\dfrac{\lambda_t^m}{\sum_{j \in \mathcal{M}_m}\lambda_j^m}, &
\text{if } \sum_{j \in \mathcal{M}_m}\lambda_j^m \geq \epsilon,\\[1em]
\dfrac{1}{|\mathcal{M}_m|}, &
\text{otherwise},
\end{cases}
\end{equation}
where $\epsilon=10^{-6}$ in all experiments. 
This fallback ensures stability when the instruction provides negligible attention to a modality. The normalization is performed within each modality rather than globally across all tokens. 
Within-modality normalization avoids bias toward streams with more tokens or different encoder statistics.

\subsection{Instruction-Driven Modality-Level Gating}
\label{subsec:method3}
\textbf{Unified Learnable Control Token.} 
We add a learnable control embedding $\mathbf{t}_c$ and insert it into the beginning of each modality token sequence. Before the cross-modal self-attention layer, the control token first summarizes the tokens of each modality using a single cross-attention block:
\begin{equation}
\mathbf{t}_c^m = \mathrm{Attn}(\mathbf{t}_c W_Q^c, \mathbf{X}_m W_K^c, \mathbf{X}_m W_V^c),
\end{equation}
where $W_Q^c$, $W_K^c$, and $W_V^c$ are lightweight projection matrices and $\mathbf{X}_m$ denotes the projected tokens of modality $m$. The resulting modality-specific control token is inserted before the tokens of that stream, as shown in Fig.~\ref{fig:overview}(b). Because the same base embedding is reused across streams, the model obtains a common interface for comparing modality summaries rather than learning unrelated gating heads for each modality's input. This choice is especially useful when some modalities are absent, because the available streams can still be scored with the same mechanism.

\noindent\textbf{Modality-Level Gating.}
Using the same cross-modal attention map $\mathbf{A}$ from Eq.~\ref{eq:attn}, we compute a scalar importance weight for each modality from the attention between instruction tokens $\mathcal{I}$ and the modality-specific control token $\mathbf{t}_c^m$. For modality $m$, we define:
\begin{equation}
\eta_m = \frac{1}{n_h}\sum_{h=1}^{n_h} \sum_{i \in \mathcal{I}} \mathbf{A}_{i \rightarrow \mathbf{t}_c^m}^{(h)}.
\end{equation}
We then normalize these scores across all $N_m$ modalities to obtain the modality-level gating coefficients:
\begin{equation}
\beta_m = \frac{\eta_m}{\sum_{m'=1}^{N_m} \eta_{m'}}.
\end{equation}
If $\sum_{m'=1}^{N_m}\eta_{m'} < 10^{-6}$, we set $\beta_m=1/N_m$ for all available modalities, mirroring the inner-modality fallback when the instruction assigns negligible attention to all control tokens.
Finally, the inner-modality weights and modality-level coefficients are combined to produce the routed representation for modality $m$:
\begin{equation}
\label{eq:gated_representation}
\hat{\mathbf{H}}_m = \mathbf{O}_m + \beta_m \, (\mathbf{w}_m \odot \mathbf{O}_m),
\end{equation}
where $\mathbf{O}_m$ denotes the self-attention output content tokens of modality $m$, and $\odot$ denotes token-wise reweighting with broadcasting over the feature dimension.
The gated modality tokens replace the corresponding modality positions before the subsequent residual, normalization, and feed-forward update. Intuitively, $\beta_m$ adjusts the overall contribution of modality $m$, while $\mathbf{w}_m$ emphasizes instruction-relevant tokens within that modality. Because $\mathbf{w}_m$ is normalized within a modality, the gating term is deliberately conservative, serving as a relative soft reweighting signal rather than replacing the original representation. UniMVU therefore does not discard a stream when its gate is small, but increases the relative influence of evidence selected by the instruction. This reduces the risk of an early wrong routing decision while still giving the decoder a more question-aligned representation.

\subsection{Training Objective and Modality Scalability}
\label{subsec:method4}
Given an instruction $x$, the available modality set $\mathcal{A}$, and answer tokens $\mathbf{y}=(y_1,\dots,y_T)$, UniMVU is trained with the standard autoregressive language-modeling loss
\begin{equation}
\mathcal{L}_{\text{ans}} = -\sum_{t=1}^{T}\log p_{\theta}\!\left(y_t \mid y_{<t}, x, \{\mathbf{X}_m\}_{m\in\mathcal{A}}\right).
\end{equation}
Only the modality projectors, control-token projections, fusion block, and LoRA parameters in the LLM are updated; all modality encoders remain frozen. This isolates the effect of instruction-aware gating from relearning the upstream encoders and keeps the adaptation regime comparable to lightweight Video LLM tuning. During inference, the same checkpoint can be applied to any subset of supported modalities by omitting missing streams from the unified sequence and renormalizing the gates over the available set.

Compared with simple token concatenation, UniMVU adds one fusion layer over the projected multimodal tokens and one single-query control-token summarization per available stream. The control-token step scales linearly with the number of tokens in a stream, and modality-level gating adds only one scalar score per modality. 
The router therefore remains lightweight relative to LLM decoding while exposing an explicit decision surface over token- and stream-level relevance.
This property is important for unified training, where the same fusion interface generalizes across different modality combinations without introducing separate task-specific connectors.


\section{Experiments and Results}
\label{sec:experiment}

\begin{table*}[t]
\caption{Task A audio--visual QA results on Music-AVQA, AVQA, and AVSD. Music-AVQA reports modality-averaged accuracy, AVQA reports accuracy, and AVSD reports ROUGE\_L and CIDEr. The asterisk marks reproduced baselines run with released code or matched evaluation protocols; UniMVU denotes task-specific training, and UniMVU$^\dagger$ denotes unified multi-task training with a single checkpoint across all tasks. \(\Delta\) reports the gain over PAVE$^*$. \best{Bold} indicates the best and \second{underlining} indicates the second-best.}
\label{tab:taskA_single}
\centering
\footnotesize
\setlength{\tabcolsep}{4.5pt}

\begingroup
\newcommand{\hl}[1]{\cellcolor{UniMVUHL}#1}

\resizebox{\linewidth}{!}{
\begin{tabular}{llrrrrrrr}
\toprule
\multicolumn{2}{c}{} &
\multicolumn{4}{c}{\textbf{Music-AVQA}} &
\multicolumn{1}{c}{\textbf{AVQA}} &
\multicolumn{2}{c}{\textbf{AVSD}} \\
\cmidrule(lr){3-6}\cmidrule(lr){7-7}\cmidrule(l){8-9}
\textbf{Size} & \textbf{Method} &
\textbf{Audio Avg.} & \textbf{Visual Avg.} & \textbf{AV Avg.} & \textbf{Overall Avg.} &
\textbf{ACC (\%)} &
\textbf{ROUGE\_L} & \textbf{CIDEr} \\
\midrule

\multirow{9}{*}{0.5B}
& PSTP-Net~\cite{pstpnet}
& \textemdash & \textemdash & \textemdash & \textemdash & 90.2 & \textemdash & \textemdash \\
& VAST~\cite{vast}
& \textemdash & \textemdash & \textemdash & \second{80.7} & \textemdash & \textemdash & \textemdash \\
& AVAF-Net~\cite{avafnet}
& 78.1 & 82.3 & 72.1 & 75.9 & \textemdash & \textemdash & \textemdash \\
& AV-Master~\cite{avmaster}
& \best{79.9} & 86.5 & 74.2 & 78.5 & \second{91.4} & \textemdash & \textemdash \\
& LLaVA-OV-FT (video-only)~\cite{llava_onevision}
& 69.6 & 76.3 & 62.8 & 67.6 & 86.4 & \textemdash & 117.6 \\
& LLaVA-OV-FT$^*$ (video-audio concat)~\cite{llava_onevision}
& 76.2 & 89.1 & 72.4 & 77.5 & 89.9 & 35.7 & 127.8 \\
& PAVE$^*$~\cite{pave}
& 75.9 & 88.6 & 72.4 & 77.3 & 89.6 & 36.5 & 134.9 \\
& \hl{\oursmethod{UniMVU$^\dagger$}}
& \hl{77.2} & \hl{\second{90.2}} & \hl{\second{74.8}} & \hl{79.3} & \hl{91.1} & \hl{\second{37.8}} & \hl{\second{145.9}} \\
& \hl{\oursmethod{UniMVU}}
& \hl{\second{79.5}} & \hl{\best{91.8}} & \hl{\best{76.7}} & \hl{\best{81.9}} & \hl{\best{92.3}} & \hl{\best{38.2}} & \hl{\best{147.1}} \\

\midrule
& \textit{$\Delta$ over PAVE$^*$ (UniMVU$^\dagger$)}
& \textit{+1.3} & \textit{+1.6} & \textit{+2.4} & \textit{+2.0} & \textit{+1.5} & \textit{+1.3} & \textit{+11.0} \\
& \textit{$\Delta$ over PAVE$^*$ (UniMVU)}
& \textit{+3.6} & \textit{+3.2} & \textit{+4.3} & \textit{+4.6} & \textit{+2.7} & \textit{+1.7} & \textit{+12.2} \\

\midrule

\multirow{5}{*}{7B}
& CAT-FT~\cite{cat}
& \best{84.9} & 86.1 & \best{83.2} & \best{84.3} & 92.0 & \textemdash & \textemdash \\
& LLaVA-OV-FT (video-only)~\cite{llava_onevision}
& 75.4 & 89.3 & 72.3 & 77.4 & 90.8 & \textemdash & 124.9 \\
& PAVE$^*$~\cite{pave}
& 79.1 & 92.7 & 77.8 & 81.9 & \second{93.4} & 38.5 & 151.6 \\
& \hl{\oursmethod{UniMVU$^\dagger$}}
& \hl{78.9} & \hl{\second{92.8}} & \hl{77.2} & \hl{81.6} & \hl{92.2} & \hl{\second{39.5}} & \hl{\second{162.7}} \\
& \hl{\oursmethod{UniMVU}}
& \hl{\second{81.7}} & \hl{\best{93.5}} & \hl{\second{79.8}} & \hl{\second{83.7}} & \hl{\best{94.3}} & \hl{\best{39.8}} & \hl{\best{165.1}} \\

\midrule
& \textit{$\Delta$ over PAVE$^*$ (UniMVU$^\dagger$)}
& \textit{-0.2} & \textit{+0.1} & \textit{-0.6} & \textit{-0.3} & \textit{-1.2} & \textit{+1.0} & \textit{+11.1} \\
& \textit{$\Delta$ over PAVE$^*$ (UniMVU)}
& \textit{+2.6} & \textit{+0.8} & \textit{+2.0} & \textit{+1.8} & \textit{+0.9} & \textit{+1.3} & \textit{+13.5} \\

\bottomrule
\end{tabular}}
\endgroup
\end{table*}

\subsection{Experimental Setup}

We evaluate UniMVU on three task families that require different modality combinations. \textit{(i) Task A (Audio--Visual QA).} It includes Music-AVQA \cite{music-avqa}, AVSD \cite{avsd}, and AVQA \cite{avqa}, covering music, dialog, and multiple-choice settings. \textit{(ii) Task B (3D QA).} It focuses on spatial reasoning in 3D scenes using ScanQA \cite{scanqa} and SQA3D \cite{sqa3d}. \textit{(iii) Task C (Video QA).} It uses MVBench \cite{mvbench}, a multi-task video understanding benchmark for fine-grained temporal and scene reasoning.

\noindent\textbf{Tasks and datasets.}
For Task A, we train and evaluate on AVSD~\cite{avsd} (open-ended) and on AVQA~\cite{avqa} and Music-AVQA~\cite{music-avqa} (multiple-choice). AVSD contains 79k question--answer (QA) pairs across 7985 videos, and evaluation uses the AVSD@DSTC7 test split with 1000 questions. AVQA provides 40k QA pairs paired with 40k videos, and Music-AVQA comprises 32k QA pairs over 9277 videos. For Task B, ScanQA~\cite{scanqa} includes approximately 41k QA pairs over approximately 800 RGB-D indoor scans, and we evaluate on its validation set with 4675 questions. SQA3D~\cite{sqa3d} contains 650 scenes, 6.8k unique situations, 20.4k situation descriptions, and 33.4k questions; we report results on its 3519-question test set. For Task C, following PAVE~\cite{pave}, we train on a subset of LLaVA-Video-178K containing videos longer than one minute with two QA pairs per video, yielding 57k videos and 114k QA pairs, and evaluate on MVBench~\cite{mvbench}.

\noindent\textbf{Evaluation metrics.}
We use the standard metrics for each dataset: generative QA tasks (AVSD, ScanQA) use captioning metrics (CIDEr, BLEU-n, METEOR, ROUGE-L) plus exact match (EM@1); multiple-choice or classification tasks (AVQA, Music-AVQA) use accuracy; SQA3D uses EM@1 under original and refined protocols with breakdowns by question type; and MVBench uses accuracy. For MVBench, we report five representative subtasks (state change, fine-grained pose, object shuffle, action prediction, and action sequence) together with the official average over the evaluated category set.

\noindent\textbf{Implementation details.}
We compare UniMVU with PAVE (reproduced via official code) \cite{pave} and LLaVA-OneVision fine-tuned with concatenated multimodal input or video-only input \cite{llava_onevision}. We follow the training pipeline of PAVE \cite{pave} with the same frozen modality encoders. For video frames, we sample 32 frames uniformly from each video and encode the RGB frames with the pretrained vision encoder from LLaVA-OneVision~\cite{llava_onevision}. 
For audio, we follow PAVE to use ImageBind~\cite{imagebind} to extract synchronized audio features from fixed-length soundtrack windows.
For 3D QA, we use the frozen LLaVA-3D feature extractor~\cite{llava3d} and treat RGB/video tokens and 3D scene tokens as separate indexed streams. For VideoQA, we add dense temporal video features sampled at 2 fps with a \(2\times2\) spatial grid as an auxiliary stream, routed separately from the sparsely sampled RGB/video tokens. We train the modality projectors, fusion module, and LoRA adapters attached to the LLM, while keeping the modality encoders frozen and updating the base LLM only through LoRA.

\begin{table*}[t]
\caption{Task B 3D QA results on the ScanQA validation set and SQA3D test set. We report EM@1, BLEU-4, METEOR, ROUGE\_L, and CIDEr for ScanQA, and EM@1 with ``What'', ``Is'', and ``How'' question-type breakdowns for SQA3D. Values in parentheses are refined scores after answer normalization and are highlighted in gray. The asterisk marks reproduced baselines run with released code or matched evaluation protocols; UniMVU denotes task-specific training, and UniMVU$^\dagger$ denotes unified multi-task training with a single checkpoint across all tasks. \(\Delta\) reports the gain over PAVE$^*$. \best{Bold} indicates the best and \second{underlining} indicates the second-best.}
\label{tab:taskB_single_combined}
\centering
\footnotesize
\setlength{\tabcolsep}{1.2pt}

\begingroup
\providecommand{\NA}{\textemdash}
\newcommand{\oref}[2]{#1~(\begingroup\setlength{\fboxsep}{0.5pt}\colorbox{gray!18}{\strut #2}\endgroup)}
\newcommand{\hl}[1]{\cellcolor{UniMVUHL}#1}

\resizebox{\linewidth}{!}{
\begin{tabular}{ll*{9}{r}}
\toprule
\multicolumn{2}{c}{} &
\multicolumn{5}{c}{\textbf{ScanQA}} &
\multicolumn{4}{c}{\textbf{SQA3D}} \\
\cmidrule(lr){3-7}\cmidrule(l){8-11}
\textbf{Size} & \textbf{Method} &
\textbf{EM@1} & \textbf{BLEU-4} & \textbf{METEOR} & \textbf{ROUGE\_L} & \textbf{CIDEr} &
\textbf{EM@1} & \textbf{What} & \textbf{Is} & \textbf{How} \\
\midrule

\multirow{7}{*}{0.5B}
& SceSU~\cite{scesu}
& 25.1 & 13.2 & 14.9 & 35.5 & 69.6 & 46.8 & 32.2 & 64.9 & 46.2 \\
& DSPNet~\cite{dspnet}
& \best{26.5} & \best{15.4} & 15.7 & 39.3 & 78.1 & 50.4 & 38.2 & \second{66.0} & 51.2 \\
& LLaVA-OV-FT (video-only)~\cite{llava_onevision}
& \oref{20.5}{36.3} & 6.5 & 14.3 & 36.9 & 70.5 & \oref{44.1}{45.7} & \NA & \NA & \NA \\
& LLaVA-OV-FT (video-3d concat)~\cite{llava_onevision}
& \oref{10.2}{24.3} & 4.9 & 7.4 & 20.2 & 34.9 & \oref{\NA}{\NA} & \NA & \NA & \NA \\
& PAVE$^*$~\cite{pave}
& \oref{23.5}{40.4} & 12.7 & 17.1 & 42.7 & 84.9
& \oref{48.5}{50.6} & \oref{37.5}{41.6} & \oref{61.0}{62.1} & \oref{50.1}{50.3} \\
& \hl{\oursmethod{UniMVU$^\dagger$}}
& \hl{\oref{24.7}{\second{42.0}}} & \hl{\second{14.5}} & \hl{\second{17.9}} & \hl{\second{44.2}} & \hl{\second{89.7}}
& \hl{\oref{\second{50.8}}{\second{52.6}}} & \hl{\oref{\second{40.9}}{\second{44.4}}} & \hl{\oref{63.7}{\second{64.7}}} & \hl{\oref{\second{52.5}}{\second{52.7}}} \\
& \hl{\oursmethod{UniMVU}}
& \hl{\oref{\second{25.9}}{\best{43.2}}} & \hl{13.5} & \hl{\best{18.0}} & \hl{\best{44.7}} & \hl{\best{90.9}}
& \hl{\oref{\best{55.2}}{\best{57.1}}} & \hl{\oref{\best{46.5}}{\best{50.6}}} & \hl{\oref{\best{67.6}}{\best{68.4}}} & \hl{\oref{\best{57.4}}{\best{57.6}}} \\

\midrule
& \textit{$\Delta$ over PAVE$^*$ (UniMVU$^\dagger$)}
& \textit{\oref{+1.2}{+1.6}} & \textit{+1.8} & \textit{+0.8} & \textit{+1.5} & \textit{+4.8}
& \textit{\oref{+2.3}{+2.0}} & \textit{\oref{+3.4}{+2.8}} & \textit{\oref{+2.7}{+2.6}} & \textit{\oref{+2.4}{+2.4}} \\
& \textit{$\Delta$ over PAVE$^*$ (UniMVU)}
& \textit{\oref{+2.4}{+2.8}} & \textit{+0.8} & \textit{+0.9} & \textit{+2.0} & \textit{+6.0}
& \textit{\oref{+6.7}{+6.5}} & \textit{\oref{+9.0}{+9.0}} & \textit{\oref{+6.6}{+6.3}} & \textit{\oref{+7.3}{+7.3}} \\

\midrule

\multirow{6}{*}{7B}
& LLaVA-3D-7B~\cite{llava3d}
& 27.0 (45.0) & 14.5 & \best{20.7} & \best{50.1} & 91.7 & 55.6 (57.6) & \NA & \NA & \NA \\
& Scene-LLM-7B~\cite{scenellm}
& 27.2 & 12.0 & 16.6 & 40.0 & 80.0 & 54.2 & \NA & \NA & \NA \\
& LLaVA-OV-FT (video-only)~\cite{llava_onevision}
& \oref{27.4}{46.3} & \NA & 13.5 & 47.4 & 95.1 & \oref{55.8}{58.1} & \NA & \NA & \NA \\
& PAVE$^*$~\cite{pave}
& \oref{28.9}{\second{48.2}} & \second{16.0} & 19.8 & 48.8 & 102.4
& \oref{57.6}{59.9} & \oref{\second{52.3}}{\second{56.9}} & \oref{69.2}{69.9} & \oref{\second{56.3}}{\second{57.4}} \\
& \hl{\oursmethod{UniMVU$^\dagger$}}
& \hl{\oref{\second{29.2}}{\best{48.8}}} & \hl{\best{17.8}} & \hl{\second{20.1}} & \hl{\second{49.0}} & \hl{\best{104.2}}
& \hl{\oref{\second{58.1}}{\second{60.4}}} & \hl{\oref{\second{52.3}}{56.6}} & \hl{\oref{\second{70.7}}{\second{71.8}}} & \hl{\oref{\best{60.4}}{\best{61.1}}} \\
& \hl{\oursmethod{UniMVU}}
& \hl{\oref{\best{29.6}}{\best{48.8}}} & \hl{\second{16.0}} & \hl{19.8} & \hl{\second{49.0}} & \hl{\second{102.7}}
& \hl{\oref{\best{59.4}}{\best{61.6}}} & \hl{\oref{\best{53.4}}{\best{57.7}}} & \hl{\oref{\best{75.9}}{\best{76.8}}} & \hl{\oref{55.9}{56.1}} \\

\midrule
& \textit{$\Delta$ over PAVE$^*$ (UniMVU$^\dagger$)}
& \textit{\oref{+0.3}{+0.6}} & \textit{+1.8} & \textit{+0.3} & \textit{+0.2} & \textit{+1.8}
& \textit{\oref{+0.5}{+0.5}} & \textit{\oref{+0.0}{-0.3}} & \textit{\oref{+1.5}{+1.9}} & \textit{\oref{+4.1}{+3.7}} \\
& \textit{$\Delta$ over PAVE$^*$ (UniMVU)}
& \textit{\oref{+0.7}{+0.6}} & \textit{+0.0} & \textit{+0.0} & \textit{+0.2} & \textit{+0.3}
& \textit{\oref{+1.8}{+1.7}} & \textit{\oref{+1.1}{+0.8}} & \textit{\oref{+6.7}{+6.9}} & \textit{\oref{-0.4}{-1.3}} \\

\bottomrule
\end{tabular}}
\endgroup
\end{table*}

\noindent\textbf{Training regimes.}
For task-specific training, separate UniMVU instances are fine-tuned on individual datasets with one cross-modal self-attention layer and gating modules; the hidden dimension follows the corresponding LLM, and the fusion layer uses 14 attention heads. For LoRA layers in the LLM, we use LoRA rank \(r=64\) and \(\alpha=128\) for 0.5B models, and \(r=128\) and \(\alpha=256\) for 7B models. Learning rates are selected from \(1\times10^{-5}\) to \(6\times10^{-5}\) depending on the dataset, and the warmup ratio is 0.03. Unless noted otherwise, we use AdamW, a cosine schedule with linear warmup, and early stopping on per-task validation. All modality encoders remain frozen, while the projectors, fusion block, and LoRA adapters are trainable. This setup isolates the effect of the gating module from full encoder adaptation and keeps the training cost comparable to lightweight Video LLM adaptation.

\noindent\textbf{Unified training protocol.}
We also train a single unified multi-task UniMVU model over all datasets. Each mini-batch is sampled from a mixed, shuffled pool using \(\alpha\)-sampling, where the probability for dataset \(i\) is proportional to \(N_i^\alpha\), and we use \(\alpha=0.5\) to reduce dominance by larger datasets. The unified model shares the same projectors, fusion block, and LoRA adapters across tasks; absent modalities are masked out and are not included in the gate normalization. This protocol tests whether one gating interface can support different modality combinations rather than whether a separately tuned model can be optimized for each benchmark.

\noindent\textbf{Evaluation and reproducibility.}
We follow the official splits and metrics of each benchmark whenever available. For open-ended QA, generated answers are post-processed with the same normalization used by the corresponding benchmark or baseline implementation before computing EM@1, BLEU-4, METEOR, ROUGE\_L, and CIDEr. For multiple-choice VideoQA, we parse the generated option or answer text and use the official MVBench categories. Baseline numbers marked with an asterisk are reproduced with the corresponding released code or evaluation protocol under the same backbone scale when possible. Because published systems often differ in backbones, modalities, and evaluation protocols, we treat reproduced PAVE and LLaVA-OV variants as the primary controlled baselines; other published results provide contextual comparisons. All reported UniMVU results use the same inference prompts within each dataset family, and the model checkpoint is selected by validation performance rather than test-set tuning.

\noindent\textbf{Compute.}
Our UniMVU runs and reproduced baselines use 2 NVIDIA H200 GPUs. For separate training, the per-dataset times for 0.5B/7B models are: Music-AVQA, 2 epochs (12 h/13 h); AVQA, 1 epoch (13 h/14 h); AVSD, 2 epochs (7 h/8 h); SQA3D, 2 epochs (13 h/14 h); ScanQA, 2 epochs (12 h/14 h); and LLaVA-Video-178K subset, 1 epoch (23 h/28 h), totaling 80h for 0.5B and 91h for 7B. Unified multi-task training completes in 81 h for the 0.5B model and 93 h for the 7B model on the same hardware, yielding one checkpoint for all evaluated task families rather than separate task-specific checkpoints.

\subsection{Task A: Audio--Visual QA}
\label{subsec:task_a}

\begin{table}[t]
\captionsetup{skip=8pt}
\caption{Task C VideoQA results on MVBench with 0.5B and 7B models.
We report five representative subtasks: state change (SC), fine-grained pose (FGP), object shuffle (OS), action prediction (AP), and action sequence (AS).
Avg. denotes the average over all 20 category sets and therefore may not equal the mean of the displayed subtasks. The asterisk marks reproduced baselines run with released code or matched evaluation protocols; UniMVU denotes task-specific training, and UniMVU$^\dagger$ denotes unified multi-task training with a single checkpoint across all tasks. \(\Delta\) reports the gain over PAVE$^*$. \best{Bold} indicates the best and \second{underlining} indicates the second-best.}
\label{tab:taskC_mvbench_combined}
\centering
\footnotesize
\setlength{\tabcolsep}{10pt}
\newcommand{\hl}[1]{\cellcolor{UniMVUHL}#1}
\begin{tabular}{llrrrrrr}
\toprule
\multicolumn{2}{c}{} & \multicolumn{6}{c}{\textbf{MVBench}} \\
\cmidrule(l){3-8}
\textbf{Size} & \textbf{Method} &
\textbf{SC} & \textbf{FGP} & \textbf{OS} & \textbf{AP} & \textbf{AS} & \textbf{Avg.} \\
\midrule

\multirow{4}{*}{0.5B}
& LLaVA-OV~\cite{llava_onevision}
& 37.5 & 49.0 & \best{33.0} & \textemdash & \textemdash & 45.5 \\
& PAVE$^*$~\cite{pave}
& \second{41.0} & \second{50.0} & \second{32.0} & 43.0 & 46.5 & 44.5 \\
& \hl{\oursmethod{UniMVU$^\dagger$}}
& \hl{37.5} & \hl{49.0} & \hl{30.5} & \hl{\best{64.0}} & \hl{\best{63.0}} & \hl{\best{48.6}} \\
& \hl{\oursmethod{UniMVU}}
& \hl{\best{43.0}} & \hl{\best{50.5}} & \hl{30.0} & \hl{\second{53.5}} & \hl{\second{52.0}} & \hl{\second{46.7}} \\

\midrule
& \textit{$\Delta$ over PAVE$^*$ (UniMVU$^\dagger$)}
& \textit{$-3.5$} & \textit{$-1.0$} & \textit{$-1.5$} & \textit{+21.0} & \textit{+16.5} & \textit{+4.1} \\
& \textit{$\Delta$ over PAVE$^*$ (UniMVU)}
& \textit{+2.0} & \textit{+0.5} & \textit{$-2.0$} & \textit{+10.5} & \textit{+5.5} & \textit{+2.2} \\

\midrule

\multirow{6}{*}{7B}
& VideoChat2-7B
& 44.0 & 49.0 & \best{42.5} & 47.5 & 66.0 & 51.1 \\
& VideoLLaMA2.1-7B
& \textemdash & \textemdash & \textemdash & \textemdash & \textemdash & 57.3 \\
& PAVE$^*$
& 51.0 & 53.5 & \second{39.5} & 70.5 & 70.7 & 57.1 \\
& LLaVA-OV-7B
& \best{52.0} & 53.0 & 35.5 & \textemdash & \textemdash & 56.7 \\
& \hl{\oursmethod{UniMVU$^\dagger$}}
& \hl{\second{51.5}} & \hl{\best{58.0}} & \hl{\second{39.5}} & \hl{\best{76.5}} & \hl{\second{76.1}} & \hl{\best{59.5}} \\
& \hl{\oursmethod{UniMVU}}
& \hl{51.0} & \hl{\second{54.5}} & \hl{38.5} & \hl{\second{71.0}} & \hl{\best{77.0}} & \hl{\second{58.0}} \\

\midrule
& \textit{$\Delta$ over PAVE$^*$ (UniMVU$^\dagger$)}
& \textit{+0.5} & \textit{+4.5} & \textit{+0.0} & \textit{+6.0} & \textit{+5.4} & \textit{+2.4} \\
& \textit{$\Delta$ over PAVE$^*$ (UniMVU)}
& \textit{+0.0} & \textit{+1.0} & \textit{$-1.0$} & \textit{+0.5} & \textit{+6.3} & \textit{+0.9} \\

\bottomrule
\end{tabular}
\end{table}

Here, we assess complementary facets of audio--visual reasoning using three public benchmarks: \textbf{Music-AVQA}, \textbf{AVSD}, and \textbf{AVQA}. 
Table~\ref{tab:taskA_single} compares task-specific UniMVU and unified UniMVU$^\dagger$ against prior methods and controlled reproduced baselines. We interpret the results along three axes: whether the model can specialize to the requested modality source, whether the fusion remains robust across dataset formats, and whether the benefit changes with LLM capacity.

First, task-specific UniMVU improves over the reproduced PAVE$^*$ baseline across all three audio--visual benchmarks and both model scales, while remaining competitive with task-specialized prior methods such as CAT-FT on Music-AVQA. At 0.5B, UniMVU improves Music-AVQA overall accuracy from 77.3 to 81.9 (+4.6), AVQA accuracy from 89.6 to 92.3 (+2.7), and AVSD CIDEr from 134.9 to 147.1 (+12.2). At 7B, the gains remain positive on the aggregate metrics, improving Music-AVQA overall accuracy from 81.9 to 83.7 (+1.8), AVQA accuracy from 93.4 to 94.3 (+0.9), and AVSD CIDEr from 151.6 to 165.1 (+13.5). The comparison is deliberately made against reproduced PAVE$^*$ results because PAVE is the closest static side-channel adaptation baseline.

The split-level Music-AVQA results show that the improvement is not produced by simply increasing the contribution of one modality for every sample. Audio, visual, and joint audio--visual questions each require a different routing behavior: acoustic questions should privilege waveform evidence, visual questions should suppress irrelevant sound, and AV questions should preserve both streams. UniMVU improves the 0.5B audio, visual, and AV averages by +3.6, +3.2, and +4.3 over PAVE$^*$, respectively, indicating that the gate adapts across question types instead of applying a single audio-heavy or video-heavy prior. The same trend is visible at 7B, where UniMVU obtains the best visual average and the second-best audio, AV, and overall averages among the compared 7B rows.

The AVQA and AVSD results further illustrate robustness across answer formats. AVQA is a multiple-choice setting in which the correct option can depend on identifying the sounding object or event; UniMVU improves over PAVE$^*$ at both scales, showing that query-conditioned gating helps even when the answer space is constrained. AVSD is open-ended and dialogue-conditioned, so surface-form overlap metrics alone may hide semantic improvements. The larger CIDEr gains relative to ROUGE\_L suggest that UniMVU better recovers content-bearing details from video/audio context, rather than merely changing short lexical patterns. This aligns with the qualitative examples in which UniMVU corrects generic or visually biased answers by shifting weight to the modality requested by the question.

The unified model UniMVU$^\dagger$ uses a single checkpoint trained jointly across all task families and still preserves strong audio--visual reasoning without task-specific fusion rules. At 0.5B, UniMVU$^\dagger$ improves over PAVE$^*$ by +2.0 overall accuracy on Music-AVQA, +1.5 accuracy on AVQA, and +11.0 CIDEr on AVSD. Its 7B audio--visual scores are more mixed, especially on Music-AVQA and AVQA, which indicates a natural trade-off between unified cross-task generalization and task-specialized optimization. Nevertheless, the strong AVSD CIDEr gain at 7B and the consistent 0.5B gains show that the shared gating interface transfers useful modality-selection behavior across the multi-task mixture.

Finally, UniMVU outperforms video-only and naive concatenation baselines, confirming that the gains are not explained by the visual backbone alone or by adding more tokens. At 0.5B, UniMVU raises Music-AVQA overall accuracy from 67.6 for LLaVA-OV-FT (video-only) to 81.9, and improves AVSD CIDEr from 117.6 to 147.1. Compared with video-audio concatenation, UniMVU also improves Music-AVQA overall accuracy from 77.5 to 81.9 and AVSD CIDEr from 127.8 to 147.1. Together with the ablations in Table~\ref{tab:ablation_baseline}, these comparisons support the claim that instruction-aware gating contributes beyond static fusion.

\subsection{Task B: 3D QA}
\label{subsec:task_b}

In this task, we evaluate grounded 3D reasoning with two complementary benchmarks, including \textbf{ScanQA} and \textbf{SQA3D}. To reflect standard evaluation practice, we show both the Original scoring and a Refined variant that normalizes minor textual mismatches. 

As shown in Table~\ref{tab:taskB_single_combined}, UniMVU improves over PAVE$^*$ on most major metrics, with especially clear gains at the 0.5B scale and on SQA3D.

At 0.5B, UniMVU lifts ScanQA EM@1 with absolute gains of 2.4 and 2.8 on the original and refined metrics, respectively. It also improves generation quality across all metrics, with a particularly large CIDEr gain of +6.0. On SQA3D, UniMVU improves EM@1 by +6.7 and +6.5 under the original and refined protocols, respectively.

The unified model UniMVU$^\dagger$ is also strong on 3D reasoning. At 0.5B, it improves over PAVE$^*$ on all ScanQA metrics, including +1.2 EM@1, +1.8 BLEU-4, +1.5 ROUGE\_L, and +4.8 CIDEr. On SQA3D, it improves overall EM@1 by +2.3 and retains gains across \textit{What}, \textit{Is}, and \textit{How} questions. 

At 7B, UniMVU$^\dagger$ is competitive with task-specific UniMVU and achieves the strongest BLEU-4, CIDEr, and \textit{How} scores among the PAVE/UniMVU variants, while remaining close on the remaining metrics.

The SQA3D gains, especially on \textit{What} and \textit{How} questions at 0.5B, are consistent with the view that instruction-aware gating helps the model use RGB/3D evidence for spatial relations and object interactions.
On ScanQA, the CIDEr gain is larger than the BLEU gain, suggesting better agreement with content-heavy references rather than only surface-form normalization. The original and refined metrics further show that the improvements remain after answer normalization, so the gains are not driven by minor textual variants. The larger margins at 0.5B than at 7B (e.g., +6.0 CIDEr and +6.7 EM@1) suggest that gating is especially helpful when the decoder is less able to absorb auxiliary 3D evidence by capacity alone.

\begin{table}[!t]
\captionsetup{skip=8pt}
\caption{Architectural component ablation on Music-AVQA. Adding cross-modal self-attention, inner-modality gating, and modality-level gating progressively improves performance over the concatenation baseline on this benchmark. The shaded row marks the final UniMVU design.}
\label{tab:ablation_baseline}
\centering
\footnotesize
\setlength{\tabcolsep}{10pt}
\renewcommand{\arraystretch}{1.3}
\begin{tabular}{lrrrr}
\toprule
\textbf{Method} &
\textbf{Audio} & \textbf{Visual} & \textbf{AV} & \textbf{Overall} \\
\midrule
LLaVA-OV-0.5B-FT-CONCAT 
& 76.2 & 89.1 & 72.4 & 77.5 \\
+ Cross-modal Self-Attention Layer
& 76.8 & 90.0 & 74.1 & 78.8 \\
+ Inner-modality Gating 
& 77.1 & 90.2 & 76.1 & 80.4 \\
\rowcolor{UniMVUHL}
\oursmethod{+ Modality-level Gating (UniMVU)} 
& \best{79.5} & \best{91.8} & \best{76.7} & \best{81.9} \\
\bottomrule
\end{tabular}
\end{table}

\subsection{Task C: VideoQA}
\label{subsec:task_c}

This task evaluates question answering that depends on visual dynamics rather than static appearance. 
This setting tests whether the model can use temporally localized evidence to answer fine-grained event questions.
We evaluate \textbf{MVBench} at both 0.5B and 7B scales, reporting subtask scores in Table~\ref{tab:taskC_mvbench_combined}. At 0.5B, task-specific UniMVU improves the average score from 44.5 to 46.7 over PAVE$^*$, while the unified model UniMVU$^\dagger$ further improves the average to 48.6. The largest gains appear on temporally demanding subtasks, including AP and AS, where UniMVU$^\dagger$ improves over PAVE$^*$ by +21.0 and +16.5, respectively. At 7B, task-specific UniMVU improves the average score from 57.1 to 58.0, and UniMVU$^\dagger$ further increases it to 59.5, with strong gains on FGP, AP, and AS.

UniMVU's fusion block routes attention across complementary video streams, including global clip tokens and dense low-resolution tokens, allowing the model to emphasize temporal evidence when the query depends on motion and event order. 
This selective gating is reflected most clearly in the strong gains on AP/AS, where the answer depends on evolving motion rather than a single salient frame. The 7B results further show that the unified checkpoint improves the average score and obtains the best FGP, AP, and AS among the compared variants, demonstrating that dense temporal evidence can be shared effectively through the same gating interface. Appearance-centric subtasks such as state change, fine-grained pose, and object shuffle rely more heavily on local visual details, so their gains are naturally smaller than those of action-centric subtasks. This pattern matches UniMVU's design: when the question implies future or sequential reasoning, the router increases the contribution of dense temporal evidence; when the question is more local, the RGB stream remains the primary source of information.

\subsection{Unified Multi-Task Training}
\label{subsec:unified_training}

We further evaluate UniMVU with a single model trained jointly across all three task families on six datasets. The unified UniMVU model, denoted as UniMVU$^\dagger$, is trained on all datasets jointly for 2 epochs with \(\alpha\)-sampling, where \(\alpha=0.5\). Encoders remain frozen, while projection layers, the fusion module, and LoRA adapters are trainable. At inference time, the same checkpoint is evaluated across Tasks A--C without changing the architecture, adding task-specific fusion heads, or fine-tuning on each benchmark separately.

The unified results in Tables~\ref{tab:taskA_single},~\ref{tab:taskB_single_combined}, and~\ref{tab:taskC_mvbench_combined} show that the proposed gating interface transfers from task-specific adaptation to unified multimodal video understanding. On Task A, UniMVU$^\dagger$ keeps strong audio--visual performance and improves over PAVE$^*$ on representative aggregate metrics, especially at the 0.5B scale. On Task B, the same checkpoint improves over PAVE$^*$ on ScanQA and SQA3D, showing that the gating module can also handle RGB-D/3D evidence for spatial reasoning. On Task C, UniMVU$^\dagger$ achieves the best average MVBench score among the compared variants at both 0.5B and 7B, with particularly large gains on action prediction and action sequence.

These results validate the central claim of UniMVU: heterogeneous streams can be handled by one instruction-aware gating mechanism rather than by separate hand-designed connectors for audio--visual, 3D, and dense-video tasks. The common control-token interface gives each available stream a comparable modality summary, and the instruction-conditioned gates convert this summary into sample-specific stream weights. As a result, the unified checkpoint can emphasize audio for acoustic questions, depth or 3D cues for spatial questions, and dense-video tokens for long-horizon temporal questions while keeping a shared LLM decoder.

The unified setting is also a stronger test than training one model per dataset because it exposes the router to different answer formats, modality combinations, and evidence types. 
The gains over video-only, concatenation, and PAVE$^*$ baselines are consistent with improvements from selective instruction-aware fusion rather than simply from adding more tokens.
Together with the ablations and qualitative examples, the unified experiments demonstrate that UniMVU provides a scalable recipe for multimodal video understanding across multiple tasks.

\begin{table}[!t]
\centering
\begin{minipage}[t]{0.48\textwidth}
\centering
\captionsetup{skip=8pt}
\caption{Control-token design ablation on Music-AVQA. A single unified control token gives best overall result.}
\label{tab:ablation_special_token}
\scriptsize
\setlength{\tabcolsep}{1.8pt}
\resizebox{\linewidth}{!}{%
\begin{tabular}{lrrrr}
\toprule
\textbf{Method} &
\textbf{Audio} & \textbf{Visual} & \textbf{AV} & \textbf{Overall} \\
\midrule
modality-specific control token (num=1) 
& 75.7 & 90.8 & 75.9 & 79.2 \\
unified control token (num=8)
& 79.3 & \textbf{91.8} & 76.1 & 80.8 \\
\rowcolor{UniMVUHL}
\oursmethod{unified control token (num=1)}
& \best{79.5} & \best{91.8} & \best{76.7} & \best{81.9} \\
\bottomrule
\end{tabular}}
\end{minipage}\hfill
\begin{minipage}[t]{0.48\textwidth}
\centering
\captionsetup{skip=8pt}
\caption{Gating mechanism design ablation on Music-AVQA.}
\label{tab:ablation_gating}
\scriptsize
\setlength{\tabcolsep}{1.6pt}
\resizebox{\linewidth}{!}{%
\begin{tabular}{lrrrrr}
\toprule
\textbf{Method} & \textbf{\#Params(M)} &
\textbf{Audio} & \textbf{Visual} & \textbf{AV} & \textbf{Overall} \\
\midrule
UniMVU(MLP Score) & 52.0 & 77.2 & 91.3 & 75.8 & 80.2 \\
UniMVU(Cross-attn Score)   & 53.4 & 78.7 & 91.2 & 76.3  & 80.5 \\
\rowcolor{UniMVUHL}
\oursmethod{UniMVU}            & \best{51.8} & \best{79.5} & \best{91.8} & \best{76.7} & \best{81.9} \\
\bottomrule
\end{tabular}}
\end{minipage}
\end{table}



\begin{table}[!t]
\captionsetup{skip=8pt}
\caption{Residual and gate-form ablation with the 0.5B backbone across three datasets. Removing the base residual term in Eq.~\ref{eq:gated_representation} or replacing UniMVU's residual refinement with a Flamingo-style zero-initialized tanh gating reduces performance under the same training protocol.}
\label{tab:ablation_residual_hard_gate}
\centering
\footnotesize
\setlength{\tabcolsep}{10pt}
\begin{tabular}{lccc}
\toprule
\multirow{2}{*}{\textbf{Variant}} &
\textbf{Music-AVQA} & \textbf{AVSD} & \textbf{SQA3D} \\
& \textbf{Avg.} & \textbf{CIDEr} & \textbf{EM@1} \\
\midrule
UniMVU (w/o residual $\mathbf{O}_m$) & 79.1 & 143.4 & 52.4 \\
Flamingo-style gate & 79.8 & 145.4 & 53.0 \\
\rowcolor{UniMVUHL}
\oursmethod{UniMVU} & \best{81.9} & \best{147.1} & \best{55.2} \\
\bottomrule
\end{tabular}
\end{table}

\subsection{Ablations and Analysis}
\label{subsec:task_d}

\noindent\textbf{Component Ablation.} 
As shown in Table~\ref{tab:ablation_baseline}, we start from the baseline \texttt{LLaVA-OV-0.5B-FT-CONCAT}. Inserting a cross-modal self-attention layer yields a clear overall gain of 1.3 points, indicating that explicit cross-modal interaction helps joint audio--visual reasoning. 
When further adding inner-modality gating, the overall score rises to 80.4, with the largest incremental gain on the \emph{AV} split (+2.0 points), suggesting that refining signals within each modality improves cross-modal alignment. Finally, introducing modality-level gating pushes overall score to 81.9. 
Due to space and compute constraints, architectural ablations are reported on Music-AVQA as a representative setting with explicit audio/visual modality shifts;
Tables~\ref{tab:taskB_single_combined}--\ref{tab:taskC_mvbench_combined} provide cross-task outcome evidence, while finer component ablations for 3D and dense-video settings remain future work.

\textbf{Control-Token Design.}
Table~\ref{tab:ablation_special_token} compares different control-token design strategies. \textit{(i) Unified vs. modality-specific.} Holding the token count fixed (\texttt{num}=1), a unified control token outperforms a modality-specific one in this ablation. This suggests that a single, shared gating signal provides a cleaner interface for cross-modality fusion, avoiding competition between modality-specific tokens.
\textit{(ii) Number of control tokens.} With a unified design, using \texttt{num}=1 slightly but consistently beats \texttt{num}=8 on all metrics. Fewer control tokens can reduce sequence length and attention fragmentation, giving a modest accuracy gain at lower computational cost. Therefore, we adopt a single unified control token as default.

\textbf{Gating Mechanism Design.}
Table~\ref{tab:ablation_gating} compares three different ways to compute the gates in our two-level fusion. We test two learned variants: \textit{(i) MLP Score}, which applies a single MLP to the instruction-token embedding to produce modality-level gates, and \textit{(ii) Cross-Attn Score}, which uses a lightweight cross-attention between the modality token and the instruction tokens to derive inner-modality gates, against our parameter-free gating. Despite introducing additional parameters, the learned gates do not improve accuracy in this setting. Our parameter-free design achieves the best overall performance. These results suggest that attention-derived scores already provide a strong inductive bias for cross-stream comparison, while additional learned scoring heads can overfit dataset-level modality priors when supervision is limited.

\begin{figure*}[!t]
    \centering
    \includegraphics[width=\textwidth]{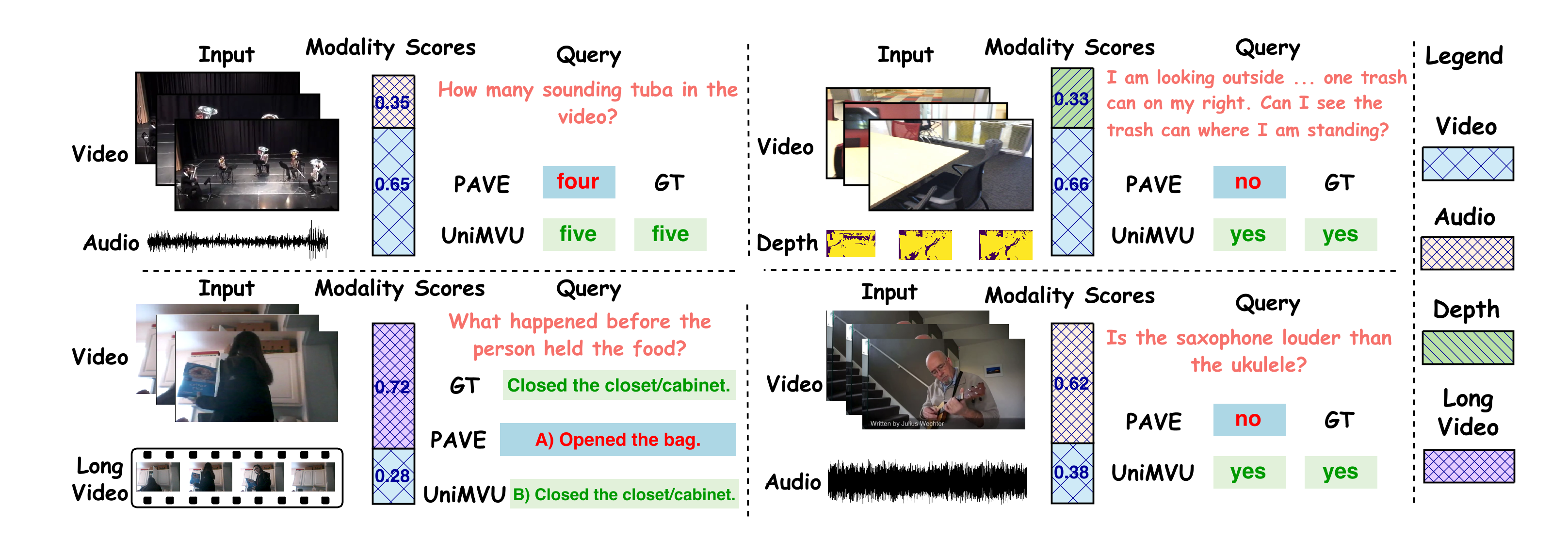}
    \caption{Qualitative comparison on representative audio-visual, 3D, and long-video examples. UniMVU adapts to the evidence required by each question and upweights video for visual counting and spatial visibility, audio for loudness comparison, and the long-video stream for temporal ordering. The right-side legend indicates each modality pattern, and the displayed scores are contextual weights produced by the gating module.}
    \label{fig:qualitative}
\end{figure*}

\begin{figure}[p]
    \centering
    \captionsetup{font=footnotesize,width=\textwidth}
    \includegraphics[width=\textwidth,height=0.6\textheight,keepaspectratio]{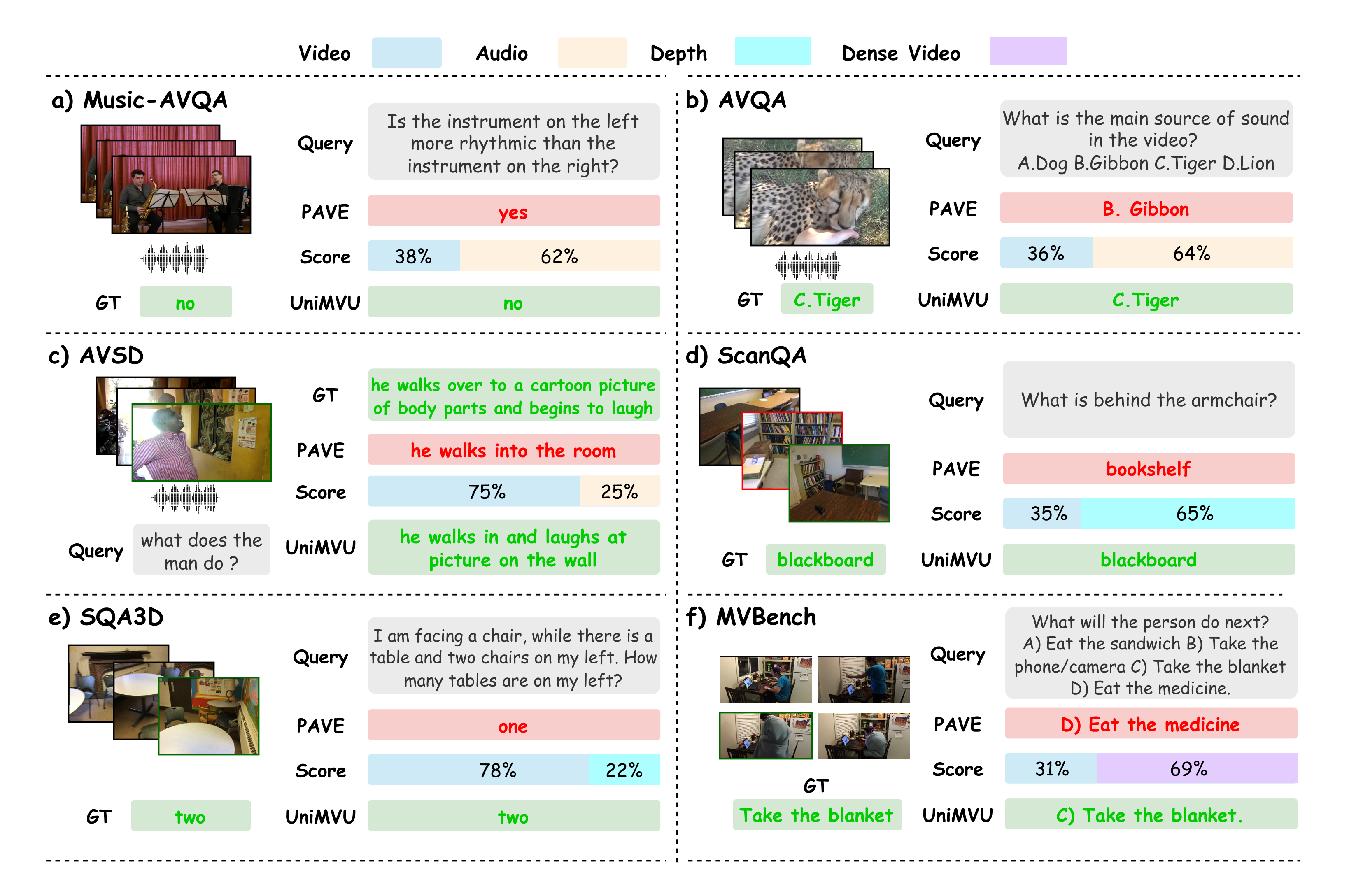}
    \caption{Additional qualitative results on three tasks across six benchmarks. Here, we show that query-conditioned modality balancing fixes PAVE’s errors. (a) \textbf{Music-AVQA}: rhythm judgment benefits from \textbf{audio} emphasis rather than its spatial understanding, and that is how UniMVU answers ``no” where PAVE says ``\textcolor{red}{yes}.”
    (b) \textbf{AVQA}: identifying the \textbf{main source of sound} prioritizes audio, letting UniMVU pick ``C. Tiger” instead of PAVE’s ``\textcolor{red}{B. Gibbon}.”
    (c) \textbf{AVSD}: describing the man’s action relies on \textbf{video} motion cues, and that is why UniMVU puts more weightage on video modality and grasps the crucial (laughs at the picture) detail.
    (d) \textbf{ScanQA}: UniMVU leans on the \textbf{depth} modality to better understand the broader scene context and locates ``blackboard” behind the armchair. Whereas PAVE gets distracted by the ``\textcolor{red}{bookshelf}" as shown with a red border in the second frame.
    (e) \textbf{SQA3D}: the green-boxed final frames reveal \emph{two} tables and UniMVU’s scoring helps count correctly where PAVE predicts ``\textcolor{red}{one}.”
    (f) \textbf{MVBench}: anticipating the next action favors the \textbf{dense-video} stream; UniMVU chooses “C) Take the blanket,” while PAVE picks an implausible alternative.}
    \label{fig:qualitative_additional}
    \vspace{0.6em}
    \includegraphics[width=\textwidth,height=0.4\textheight,keepaspectratio]{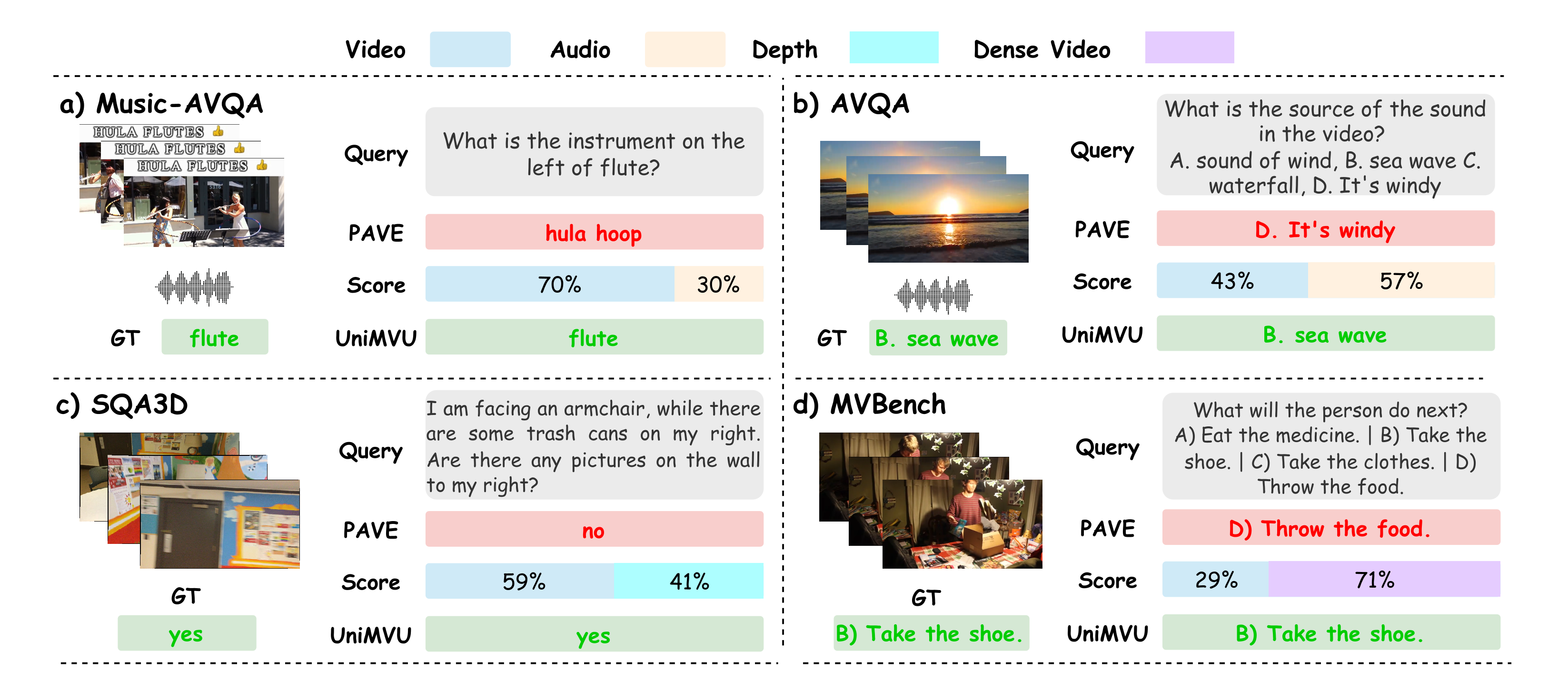}
    \caption{Additional qualitative examples across audio-visual QA, 3D QA, and VideoQA. These cases further illustrate that UniMVU changes the dominant modality according to the query instead of applying a fixed modality prior.}
    \label{fig:qualitative_additional_2}
\end{figure}

\textbf{Gate Form and Residual Path.}
Table~\ref{tab:ablation_residual_hard_gate} tests how the gating signal should be injected into the modality representation in Eq.~\ref{eq:gated_representation}. The no-residual variant removes the base $\mathbf{O}_m$ term from the gated update and degrades performance, confirming that the residual connection is critical for preserving gradient flow and base feature representations. The term $\beta_m \times (w_m \cdot O_m)$ acts as an instruction-aware \textit{adaptation}, amplifying relevant cues rather than strictly filtering them. 
The second variant replaces our gating module with a Flamingo-style gate with zero-init tanh that allows strict zero weights gating. This variant leads to lower scores than our method across benchmarks, showing the efficacy of our instruction-aware adaptation. These results support using the gate as a conservative instruction-aware adaptation that amplifies relevant evidence while preserving the attended representation, rather than as a replacement path that can overly suppress secondary cues.

\subsection{Qualitative Analysis}
\label{subsec:qualitative}


We present qualitative results in Figs.~\ref{fig:qualitative}--\ref{fig:qualitative_additional_2}. The displayed modality scores are the instruction-conditioned gating weights produced by the fusion block. In these examples, UniMVU assigns higher scores to the modality required by the question and improves over PAVE's uniform weighting. We visualize the two most important modality scores in each example to show how the gating module emphasizes video, audio, depth, or dense-video evidence.


In Fig.~\ref{fig:qualitative}, the top row shows two cases where video evidence is more important. For example, the question ``\textit{how many sounding tuba in the video?}'' depends on visual counting, and UniMVU attends more to video and predicts the correct answer, five, whereas PAVE predicts four. The bottom row shows cases where other modalities dominate. For the question ``\textit{Is the saxophone louder than the ukulele?}'', UniMVU attends more to audio and answers correctly, while PAVE fails under uniform weighting.

The additional examples in Fig.~\ref{fig:qualitative_additional} show how the attention-derived modality scores coincide with concrete corrections. 

In the \textbf{(a) Music-AVQA} example, the question ``\textit{Is the instrument on the left more rhythmic than the instrument on the right?}'' asks about rhythmicity, which is an acoustic property. UniMVU therefore assigns a higher score to audio and answers ``no,'' showing sensitivity to temporal patterns in the waveform that are not visible from appearance cues alone. By contrast, PAVE's answer ``yes'' reflects a visual bias caused by uniform modality weighting.
In \textbf{(b) AVQA}, the task is to identify the \textit{main sound source}. Here, the UniMVU again assigns more weight to audio and selects “C.\ tiger,” while PAVE’s “B.\ gibbon” reveals overreliance on what looks plausible, not what sounds correct. In \textbf{(c) AVSD}, the example requires a grounded description of an action and that is why by up–weighting the video score, UniMVU captures the micro–event (``laughs at the picture on the wall”), which PAVE misses with a generic answer (``walks into the room”). This is also consistent with our CIDEr gains that reward semantic relevance. 


Furthermore, in \textbf{(d) ScanQA}, we intentionally mark the second frame in \textcolor{red}{red} to highlight a visually dominant \emph{bookshelf}. Here, PAVE anchors to this distractor and answers “bookshelf.” UniMVU on the other hand increases the depth modality score, and effectively widens the 3D context to reason about what is behind the armchair, and recovers the correct answer. In \textbf{(e) SQA3D}, the last frames are outlined in \textcolor{green}{green} to reveal there are indeed two tables. Even with a higher video share than depth, UniMVU leverages fused visual cues to separate adjacent planar surfaces and counts each table correctly, whereas PAVE undercounts it by one. Finally, in \textbf{(f) MVBench}, anticipating the next action requires longer–horizon evidence. UniMVU increases the dense–video weight and selects “C) Take the blanket,” while PAVE’s choice suggests overfitting to a local frame. Across all panels, the absolute percentages are less important than their relative ranking. Whether audio outranks video for acoustic queries, or dense video outranks video for long–horizon reasoning, UniMVU consistently avoids salient but irrelevant distractors and grounds its prediction in the right modality, explaining the systematic corrections we observe over PAVE.

These examples complement the aggregate tables by showing that the gating scores vary at the sample level rather than following a fixed dataset prior. The same model increases audio weights for acoustic questions, depth or visual weights for spatial questions, and dense-video weights for action-anticipation questions. This sample-dependent behavior is the core advantage of UniMVU over static fusion: the modality gating source is selected according to the instruction before the LLM generates the answer.

Overall, the qualitative results support the central claim that UniMVU performs instruction-aware modality gating across multimodal video inputs. The scores are not dataset-level constants; they change with query semantics and modality availability, and the resulting answers correct representative failures of uniform weighting. This connection between sample-level gating and answer-level improvement explains the consistent gains observed in the quantitative tables.

\section{Discussion and Limitations}
\label{sec:discussion}
UniMVU demonstrates that multimodal video inputs are most effective when the model explicitly routes modality information before LLM decoding. The inner-modality gate selects instruction-relevant tokens within each modality, while the modality-level gate assigns sample-specific weights to video, audio, depth/3D, and dense temporal inputs. This design makes the fusion stage more transparent than fixed concatenation and helps explain why the gains are strongest in modality-dependent settings and lower-capacity decoders: UniMVU reduces the burden on the decoder by presenting a question-aligned multimodal representation.

\textbf{Limitations.} UniMVU currently supports video, audio, depth/3D, and dense-video streams, while broader multimodal video may also include point cloud, IMU, subtitles, multi-view geometry, and event streams. Extending to these modalities is a natural next step, but public video question-answer benchmarks with synchronized multi-sensor videos remain scarce. 
Future work can extend the same principle to additional modalities as richer synchronized datasets and standardized benchmarks become available.

\section{Conclusion}
\label{sec:conclusion}


In this paper, we present UniMVU, a unified multi-modal video understanding framework that uses context-adaptive cross-modal gating to fuse diverse inputs. Our method effectively mitigates modality interference by dynamically adjusting the contribution of each modality at both token and modality levels, enabling the model to focus on the most relevant visual, audio, or depth cues. Experiments across diverse benchmarks show consistent improvements over prior methods and state-of-the-art performance. Unified training results further demonstrate that a single gating interface can support audio-visual, 3D, and dense temporal inputs while sharing one LLM decoder. Overall, UniMVU introduces a lightweight, scalable, and interpretable approach for adaptive multimodal Video LLM fusion.

 
%


\bibliographystyle{plain}
\bibliography{main}

@String(PAMI = {IEEE Trans. Pattern Anal. Mach. Intell.})

@String(TIP = {IEEE Trans. Image Process.})

@String(TMM = {IEEE Trans. Multimedia})

@String(NIPS = {Adv. Neural Inf. Process. Syst.})

@String(CVPR = {Proc. IEEE/CVF Conf. Comput. Vis. Pattern Recognit.})

@String(ICCV = {Proc. IEEE/CVF Int. Conf. Comput. Vis.})

@String(ICCVW = {Proc. IEEE/CVF Int. Conf. Comput. Vis. Workshops})

@String(ECCV = {Proc. Eur. Conf. Comput. Vis.})

@String(WACV = {Proc. IEEE/CVF Winter Conf. Appl. Comput. Vis.})

@String(ACMMM = {Proc. ACM Int. Conf. Multimedia})

@String(ICML = {Proc. Int. Conf. Mach. Learn.})

@String(ICLR = {Proc. Int. Conf. Learn. Represent.})

@String(TMLR = {Trans. Mach. Learn. Res.})

@String(AAAI = {Proc. AAAI Conf. Artif. Intell.})

@String(ACL = {Proc. Annu. Meeting Assoc. Comput. Linguistics})

@String(EMNLPDEMO = {Proc. Conf. Empirical Methods Natural Lang. Process.: Syst. Demonstrations})

@String(ICMR = {Proc. ACM Int. Conf. Multimedia Retrieval})

@inproceedings{VideoChatGPT,
  title = {{Video-ChatGPT}: Towards Detailed Video Understanding via Large Vision and Language Models},
  author = {Maaz, Muhammad and Rasheed, Hanoona and Khan, Salman and Khan, Fahad},
  booktitle = ACL,
  pages = {12585--12602},
  year = {2024},
  doi = {10.18653/v1/2024.acl-long.679}
}

@inproceedings{flamingo,
  title = {{Flamingo}: A Visual Language Model for Few-Shot Learning},
  author = {Alayrac, Jean-Baptiste and Donahue, Jeff and Luc, Pauline and Miech, Antoine and Barr, Iain and Hasson, Yana and Lenc, Karel and Mensch, Arthur and Millican, Katie and Reynolds, Malcolm and Ring, Roman and Rutherford, Eliza and Cabi, Serkan and Han, Tengda and Gong, Zhitao and Samangooei, Sina and Monteiro, Marianne and Menick, Jacob and Borgeaud, Sebastian and Brock, Andrew and Nematzadeh, Aida and Sharifzadeh, Sahand and Binkowski, Mikolaj and Barreira, Ricardo and Vinyals, Oriol and Zisserman, Andrew and Simonyan, Karen},
  booktitle = NIPS,
  volume = {35},
  pages = {23716--23736},
  year = {2022}
}

@inproceedings{videollama,
  title = {{Video-LLaMA}: An Instruction-Tuned Audio-Visual Language Model for Video Understanding},
  author = {Zhang, Hang and Li, Xin and Bing, Lidong},
  booktitle = EMNLPDEMO,
  pages = {543--553},
  year = {2023},
  doi = {10.18653/v1/2023.emnlp-demo.49}
}

@article{videollama2,
  title = {{VideoLLaMA 2}: Advancing Spatial-Temporal Modeling and Audio Understanding in Video-{LLMs}},
  author = {Cheng, Zesen and Leng, Sicong and Zhang, Hang and Xin, Yifei and Li, Xin and Chen, Guanzheng and Zhu, Yongxin and Zhang, Wenqi and Luo, Ziyang and Zhao, Deli and Bing, Lidong},
  journal = {arXiv preprint arXiv:2406.07476},
  year = {2024}
}

@inproceedings{pave,
  title = {{PAVE}: Patching and Adapting Video Large Language Models},
  author = {Liu, Zhuoming and Li, Yiquan and Nguyen, Khoi Duc and Zhong, Yiwu and Li, Yin},
  booktitle = CVPR,
  pages = {3306--3317},
  year = {2025},
  doi = {10.1109/CVPR52734.2025.00314}
}

@inproceedings{cat,
  title = {{CAT}: Enhancing Multimodal Large Language Model to Answer Questions in Dynamic Audio-Visual Scenarios},
  author = {Ye, Qilang and Yu, Zitong and Shao, Rui and Xie, Xinyu and Torr, Philip H. S. and Cao, Xiaochun},
  booktitle = ECCV,
  pages = {146--164},
  year = {2024},
  doi = {10.1007/978-3-031-72684-2_9}
}

@inproceedings{cdViews,
  title = {{3D} Question Answering via Only {2D} Vision-Language Models},
  author = {Wang, Fengyun and Yu, Sicheng and Wu, Jiawei and Tang, Jinhui and Zhang, Hanwang and Sun, Qianru},
  booktitle = ICML,
  series = {Proc. Mach. Learn. Res.},
  volume = {267},
  pages = {65310--65325},
  year = {2025}
}

@inproceedings{avllm,
  title = {Audio-Visual {LLM} for Video Understanding},
  author = {Shu, Fangxun and Zhang, Lei and Jiang, Hao and Xie, Cihang},
  booktitle = ICCVW,
  pages = {4305--4314},
  year = {2025}
}

@article{llava_onevision,
  title = {{LLaVA-OneVision}: Easy Visual Task Transfer},
  author = {Li, Bo and Zhang, Yuanhan and Guo, Dong and Zhang, Renrui and Li, Feng and Zhang, Hao and Zhang, Kaichen and Zhang, Peiyuan and Li, Yanwei and Liu, Ziwei and Li, Chunyuan},
  journal = TMLR,
  year = {2025}
}

@inproceedings{llava3d,
  title = {{LLaVA-3D}: A Simple yet Effective Pathway to Empowering {LMMs} with {3D} Capabilities},
  author = {Zhu, Chenming and Wang, Tai and Zhang, Wenwei and Pang, Jiangmiao and Liu, Xihui},
  booktitle = ICCV,
  pages = {4295--4305},
  year = {2025},
  doi = {10.1109/ICCV51701.2025.00409}
}

@inproceedings{pointllm,
  title = {{PointLLM}: Empowering Large Language Models to Understand Point Clouds},
  author = {Xu, Runsen and Wang, Xiaolong and Wang, Tai and Chen, Yilun and Pang, Jiangmiao and Lin, Dahua},
  booktitle = ECCV,
  pages = {131--147},
  year = {2024},
  doi = {10.1007/978-3-031-72698-9_8}
}

@inproceedings{3dllm,
  title = {{3D-LLM}: Injecting the {3D} World into Large Language Models},
  author = {Hong, Yining and Zhen, Haoyu and Chen, Peihao and Zheng, Shuhong and Du, Yilun and Chen, Zhenfang and Gan, Chuang},
  booktitle = NIPS,
  volume = {36},
  year = {2023}
}

@inproceedings{owl2,
  title = {{mPLUG-Owl2}: Revolutionizing Multi-Modal Large Language Model with Modality Collaboration},
  author = {Ye, Qinghao and Xu, Haiyang and Ye, Jiabo and Yan, Ming and Hu, Anwen and Liu, Haowei and Qian, Qi and Zhang, Ji and Huang, Fei},
  booktitle = CVPR,
  pages = {13040--13051},
  year = {2024}
}

@article{llava_av_ssm,
  title={Does Audio Matter for Modern Video-LLMs and Their Benchmarks?},
  author={Kim, Geewook and Seo, Minjoon},
  journal={arXiv preprint arXiv:2509.17901},
  year={2025}
}

@inproceedings{attention,
  title = {Attention Is All You Need},
  author = {Vaswani, Ashish and Shazeer, Noam and Parmar, Niki and Uszkoreit, Jakob and Jones, Llion and Gomez, Aidan N. and Kaiser, Lukasz and Polosukhin, Illia},
  booktitle = NIPS,
  volume = {30},
  pages = {5998--6008},
  year = {2017}
}

@article{qwen2,
  title = {{Qwen2} Technical Report},
  author = {Yang, An and Yang, Baosong and Hui, Binyuan and Zheng, Bo and Yu, Bowen and Zhou, Chang and Li, Chengpeng and Li, Chengyuan and Liu, Dayiheng and Huang, Fei and Dong, Guanting and Wei, Haoran and Lin, Huan and Tang, Jialong and Wang, Jialin and Yang, Jian and Tu, Jianhong and Zhang, Jianwei and Ma, Jianxin and Yang, Jianxin and Xu, Jin and Zhou, Jingren and Bai, Jinze and He, Jinzheng and Lin, Junyang and Dang, Kai and Lu, Keming and Chen, Keqin and Yang, Kexin and Li, Mei and Xue, Mingfeng and Ni, Na and Zhang, Pei and Wang, Peng and Peng, Ru and Men, Rui and Gao, Ruize and Lin, Runji and Wang, Shijie and Bai, Shuai and Tan, Sinan and Zhu, Tianhang and Li, Tianhao and Liu, Tianyu and Ge, Wenbin and Deng, Xiaodong and Zhou, Xiaohuan and Ren, Xingzhang and Zhang, Xinyu and Wei, Xipin and Ren, Xuancheng and Liu, Xuejing and Fan, Yang and Yao, Yang and Zhang, Yichang and Wan, Yu and Chu, Yunfei and Liu, Yuqiong and Cui, Zeyu and Zhang, Zhenru and Guo, Zhifang and Fan, Zhihao},
  journal = {arXiv preprint arXiv:2407.10671},
  year = {2024}
}

@article{rope,
  title = {{RoFormer}: Enhanced Transformer with Rotary Position Embedding},
  author = {Su, Jianlin and Ahmed, Murtadha and Lu, Yu and Pan, Shengfeng and Bo, Wen and Liu, Yunfeng},
  journal = {Neurocomputing},
  volume = {568},
  pages = {127063},
  year = {2024},
  doi = {10.1016/j.neucom.2023.127063}
}

@inproceedings{music-avqa,
  title = {Learning to Answer Questions in Dynamic Audio-Visual Scenarios},
  author = {Li, Guangyao and Wei, Yake and Tian, Yapeng and Xu, Chenliang and Wen, Ji-Rong and Hu, Di},
  booktitle = CVPR,
  pages = {19108--19118},
  year = {2022}
}

@inproceedings{avqa,
  title = {{AVQA}: A Dataset for Audio-Visual Question Answering on Videos},
  author = {Yang, Pinci and Wang, Xin and Duan, Xuguang and Chen, Hong and Hou, Runze and Jin, Cong and Zhu, Wenwu},
  booktitle = ACMMM,
  pages = {3480--3491},
  year = {2022},
  doi = {10.1145/3503161.3548291}
}

@inproceedings{avsd,
  title = {Audio Visual Scene-Aware Dialog},
  author = {Alamri, Huda and Cartillier, Vincent and Das, Abhishek and Wang, Jue and Cherian, Anoop and Essa, Irfan and Batra, Dhruv and Marks, Tim K. and Hori, Chiori and Anderson, Peter and Lee, Stefan and Parikh, Devi},
  booktitle = CVPR,
  pages = {7558--7567},
  year = {2019}
}

@inproceedings{scanqa,
  title = {{ScanQA}: {3D} Question Answering for Spatial Scene Understanding},
  author = {Azuma, Daichi and Miyanishi, Taiki and Kurita, Shuhei and Kawanabe, Motoaki},
  booktitle = CVPR,
  pages = {19129--19139},
  year = {2022}
}

@inproceedings{sqa3d,
  title = {{SQA3D}: Situated Question Answering in {3D} Scenes},
  author = {Ma, Xiaojian and Yong, Silong and Zheng, Zilong and Li, Qing and Liang, Yitao and Zhu, Song-Chun and Huang, Siyuan},
  booktitle = ICLR,
  year = {2023}
}

@inproceedings{videomme,
  title = {{Video-MME}: The First-Ever Comprehensive Evaluation Benchmark of Multi-Modal {LLMs} in Video Analysis},
  author = {Fu, Chaoyou and Dai, Yuhan and Luo, Yongdong and Li, Lei and Ren, Shuhuai and Zhang, Renrui and Wang, Zihan and Zhou, Chenyu and Shen, Yunhang and Zhang, Mengdan and Chen, Peixian and Li, Yanwei and Lin, Shaohui and Zhao, Sirui and Li, Ke and Xu, Tong and Zheng, Xiawu and Chen, Enhong and Shan, Caifeng and He, Ran and Sun, Xing},
  booktitle = CVPR,
  pages = {24108--24118},
  year = {2025},
  doi = {10.1109/CVPR52734.2025.02245}
}

@inproceedings{mvbench,
  title = {{MVBench}: A Comprehensive Multi-Modal Video Understanding Benchmark},
  author = {Li, Kunchang and Wang, Yali and He, Yinan and Li, Yizhuo and Wang, Yi and Liu, Yi and Wang, Zun and Xu, Jilan and Chen, Guo and Luo, Ping and Wang, Limin and Qiao, Yu},
  booktitle = CVPR,
  pages = {22195--22206},
  year = {2024},
  doi = {10.1109/CVPR52733.2024.02095}
}

@inproceedings{imagebind,
  title = {{ImageBind}: One Embedding Space to Bind Them All},
  author = {Girdhar, Rohit and El-Nouby, Alaaeldin and Liu, Zhuang and Singh, Mannat and Alwala, Kalyan Vasudev and Joulin, Armand and Misra, Ishan},
  booktitle = CVPR,
  pages = {15180--15190},
  year = {2023}
}

@inproceedings{pstpnet,
  title = {Progressive Spatio-Temporal Perception for Audio-Visual Question Answering},
  author = {Li, Guangyao and Hou, Wenxuan and Hu, Di},
  booktitle = ACMMM,
  pages = {7808--7816},
  year = {2023},
  doi = {10.1145/3581783.3612293}
}

@inproceedings{languagebind,

  title = {{LanguageBind}: Extending Video-Language Pretraining to {N}-Modality by Language-Based Semantic Alignment},
  author = {Zhu, Bin and Lin, Bin and Ning, Munan and Yan, Yang and Cui, Jiaxi and Wang, HongFa and Pang, Yatian and Jiang, Wenhao and Zhang, Junwu and Li, Zongwei and Zhang, Cai and Li, Zhifeng and Liu, Wei and Yuan, Li},
  booktitle = ICLR,
  year = {2024}

}

@inproceedings{vast,
  title = {{VAST}: A Vision-Audio-Subtitle-Text Omni-Modality Foundation Model and Dataset},
  author = {Chen, Sihan and Li, Handong and Wang, Qunbo and Zhao, Zijia and Sun, Mingzhen and Zhu, Xinxin and Liu, Jing},
  booktitle = NIPS,
  volume = {36},
  pages = {72842--72866},
  year = {2023}
}

@inproceedings{avafnet,
  title = {Audio-Visual Adaptive Fusion Network for Question Answering Based on Contrastive Learning},
  author = {Zhao, Xujian and Wang, Yixin and Jin, Peiquan},
  booktitle = AAAI,
  volume = {39},
  number = {10},
  pages = {10483--10491},
  year = {2025},
  doi = {10.1609/aaai.v39i10.33138}
}

@article{avmaster,
  title = {{AV-Master}: Dual-Path Comprehensive Perception Makes Better Audio-Visual Question Answering},
  author = {Zhang, Jiayu and Ye, Shuo and Ye, Qilang and Lin, Xun and Song, Zihan and Yu, Zitong},
  journal = {arXiv preprint arXiv:2510.18346},
  year = {2025}
}

@inproceedings{scenellm,
  title = {{Scene-LLM}: Extending Language Model for {3D} Visual Reasoning},
  author = {Fu, Rao and Liu, Jingyu and Chen, Xilun and Nie, Yixin and Xiong, Wenhan},
  booktitle = WACV,
  pages = {2195--2206},
  year = {2025}
}

@inproceedings{scesu,
  title = {Scene-Guided Attention Network for Spatial Understanding in {3D} Scenes},
  author = {Jiang, Yunqi and Zhang, Jianwei and Lin, Chaoyang and Yu, Yi and Yang, Zhenguo},
  booktitle = ICMR,
  pages = {616--624},
  year = {2025},
  doi = {10.1145/3731715.3733426}
}

@inproceedings{dspnet,
  title = {{DSPNet}: Dual-Vision Scene Perception for Robust {3D} Question Answering},
  author = {Luo, Jingzhou and Liu, Yang and Chen, Weixing and Li, Zhen and Wang, Yaowei and Li, Guanbin and Lin, Liang},
  booktitle = CVPR,
  pages = {14169--14178},
  year = {2025},
  doi = {10.1109/CVPR52734.2025.01322}
}

@inproceedings{gpt3,
  title = {Language Models Are Few-Shot Learners},
  author = {Brown, Tom B. and Mann, Benjamin and Ryder, Nick and Subbiah, Melanie and Kaplan, Jared and Dhariwal, Prafulla and Neelakantan, Arvind and Shyam, Pranav and Sastry, Girish and Askell, Amanda and Agarwal, Sandhini and Herbert-Voss, Ariel and Krueger, Gretchen and Henighan, Tom and Child, Rewon and Ramesh, Aditya and Ziegler, Daniel M. and Wu, Jeffrey and Winter, Clemens and Hesse, Christopher and Chen, Mark and Sigler, Eric and Litwin, Mateusz and Gray, Scott and Chess, Benjamin and Clark, Jack and Berner, Christopher and McCandlish, Sam and Radford, Alec and Sutskever, Ilya and Amodei, Dario},
  booktitle = NIPS,
  volume = {33},
  pages = {1877--1901},
  year = {2020}
}

@article{flant5,
  title = {Scaling Instruction-Finetuned Language Models},
  author = {Chung, Hyung Won and Hou, Le and Longpre, Shayne and Zoph, Barret and Tay, Yi and Fedus, William and Li, Yunxuan and Wang, Xuezhi and Dehghani, Mostafa and Brahma, Siddhartha and Webson, Albert and Gu, Shixiang Shane and Dai, Zhuyun and Suzgun, Mirac and Chen, Xinyun and Chowdhery, Aakanksha and Castro-Ros, Alex and Pellat, Marie and Robinson, Kevin and Valter, Dasha and Narang, Sharan and Mishra, Gaurav and Yu, Adams and Zhao, Vincent and Huang, Yanping and Dai, Andrew and Yu, Hongkun and Petrov, Slav and Chi, Ed H. and Dean, Jeff and Devlin, Jacob and Roberts, Adam and Zhou, Denny and Le, Quoc V. and Wei, Jason},
  journal = {J. Mach. Learn. Res.},
  volume = {25},
  number = {70},
  pages = {1--53},
  year = {2024}
}

@article{llama,
  title = {{LLaMA}: Open and Efficient Foundation Language Models},
  author = {Touvron, Hugo and Lavril, Thibaut and Izacard, Gautier and Martinet, Xavier and Lachaux, Marie-Anne and Lacroix, Timothee and Roziere, Baptiste and Goyal, Naman and Hambro, Eric and Azhar, Faisal and Rodriguez, Aurelien and Joulin, Armand and Grave, Edouard and Lample, Guillaume},
  journal = {arXiv preprint arXiv:2302.13971},
  year = {2023}
}

@article{moviechat_plus_tpami,
  title = {{MovieChat+}: Question-Aware Sparse Memory for Long Video Question Answering},
  author = {Song, Enxin and Chai, Wenhao and Ye, Tian and Hwang, Jenq-Neng and Li, Xi and Wang, Gaoang},
  journal = PAMI,
  volume = {48},
  number = {1},
  pages = {374--389},
  year = {2026},
  doi = {10.1109/TPAMI.2025.3604614}
}

@article{xue2017unifying,
  title = {Unifying the Video and Question Attentions for Open-Ended Video Question Answering},
  author = {Xue, Hongyang and Zhao, Zhou and Cai, Deng},
  journal = TIP,
  volume = {26},
  number = {12},
  pages = {5656--5666},
  year = {2017},
  doi = {10.1109/TIP.2017.2746267}
}

@article{webvideo_vqa_tpami,
  title = {Learning to Answer Visual Questions from Web Videos},
  author = {Yang, Antoine and Miech, Antoine and Sivic, Josef and Laptev, Ivan and Schmid, Cordelia},
  journal = PAMI,
  volume = {47},
  number = {5},
  pages = {3202--3218},
  year = {2025},
  doi = {10.1109/TPAMI.2022.3173208}
}

@article{paa_videoqa_tpami,
  title = {Parse, Align and Aggregate: Graph-Driven Compositional Reasoning for Video Question Answering},
  author = {Li, Jiangtong and Liao, Zhaohe and Xiao, Fengshun and Li, Tianjiao and Zhang, Qiang and Zhao, Haohua and Niu, Li and Chen, Guang and Zhang, Liqing and Jiang, Changjun},
  journal = PAMI,
  volume = {48},
  number = {5},
  pages = {5586--5603},
  year = {2026},
  doi = {10.1109/TPAMI.2026.3650864}
}

@article{mranet_tpami,
  title = {{MRA-Net}: Improving {VQA} via Multi-Modal Relation Attention Network},
  author = {Peng, Liang and Yang, Yang and Wang, Zheng and Huang, Zi and Shen, Heng Tao},
  journal = PAMI,
  volume = {44},
  number = {1},
  pages = {318--329},
  year = {2022},
  doi = {10.1109/TPAMI.2020.3004830}
}

@article{robustvqa_tpami,
  title = {Robust Visual Question Answering: Datasets, Methods, and Future Challenges},
  author = {Ma, Jie and Wang, Pinghui and Kong, Dechen and Wang, Zewei and Liu, Jun and Pei, Hongbin and Zhao, Junzhou},
  journal = PAMI,
  volume = {46},
  number = {8},
  pages = {5575--5594},
  year = {2024},
  doi = {10.1109/TPAMI.2024.3366154}
}

@inproceedings{videosalmonn,
  title = {{video-SALMONN}: Speech-Enhanced Audio-Visual Large Language Models},
  author = {Sun, Guangzhi and Yu, Wenyi and Tang, Changli and Chen, Xianzhao and Tan, Tian and Li, Wei and Lu, Lu and Ma, Zejun and Wang, Yuxuan and Zhang, Chao},
  booktitle = ICML,
  series = {Proc. Mach. Learn. Res.},
  volume = {235},
  pages = {47198--47217},
  publisher = {PMLR},
  year = {2024}
}

@article{videosalmonn2,
  title = {{video-SALMONN 2}: Caption-Enhanced Audio-Visual Large Language Models},
  author = {Tang, Changli and Li, Yixuan and Yang, Yudong and Zhuang, Jimin and Sun, Guangzhi and Li, Wei and Ma, Zejun and Zhang, Chao},
  journal = {arXiv preprint arXiv:2506.15220},
  year = {2025},
  doi = {10.48550/arXiv.2506.15220},
  eprint = {2506.15220},
  archivePrefix = {arXiv}
}

@article{threeur_llm,
  title = {{3UR-LLM}: An End-to-End Multimodal Large Language Model for {3D} Scene Understanding},
  author = {Xiong, Haomiao and Zhuge, Yunzhi and Zhu, Jiawen and Zhang, Lu and Lu, Huchuan},
  journal = TMM,
  year = {2025},
  doi = {10.1109/TMM.2025.3557620}
}

@inproceedings{longvideobench,
  title = {{LongVideoBench}: A Benchmark for Long-Context Interleaved Video-Language Understanding},
  author = {Wu, Haoning and Li, Dongxu and Chen, Bei and Li, Junnan},
  booktitle = NIPS,
  volume = {37},
  pages = {28828--28857},
  year = {2024},
  doi = {10.52202/079017-0907}
}

@inproceedings{mlvu,
  title = {{MLVU}: Benchmarking Multi-Task Long Video Understanding},
  author = {Zhou, Junjie and Shu, Yan and Zhao, Bo and Wu, Boya and Liang, Zhengyang and Xiao, Shitao and Qin, Minghao and Yang, Xi and Xiong, Yongping and Zhang, Bo and Huang, Tiejun and Liu, Zheng},
  booktitle = CVPR,
  pages = {13691--13701},
  year = {2025},
  doi = {10.1109/CVPR52734.2025.01278}
}

@article{diagnosing,
  title = {Diagnosing and Mitigating Modality Interference in Multimodal Large Language Models},
  author = {Cai, Rui and Li, Bangzheng and Wen, Xiaofei and Chen, Muhao and Zhao, Zhe},
  journal = {arXiv preprint arXiv:2505.19616},
  year = {2025}
}

@article{catplus_tpami2025,
  title = {{CAT+}: Investigating and Enhancing Audio-Visual Understanding in Large Language Models},
  author = {Ye, Qilang and Yu, Zitong and Shao, Rui and Cui, Yawen and Kang, Xiangui and Liu, Xin and Torr, Philip H. S. and Cao, Xiaochun},
  journal = PAMI,
  volume = {47},
  number = {10},
  pages = {8674--8690},
  year = {2025},
  doi = {10.1109/TPAMI.2025.3582389}
}

@article{unimoe_tpami2025,
  title = {{Uni-MoE}: Scaling Unified Multimodal {LLMs} with Mixture of Experts},
  author = {Li, Yunxin and Jiang, Shenyuan and Hu, Baotian and Wang, Longyue and Zhong, Wanqi and Luo, Wenhan and Ma, Lin and Zhang, Min},
  journal = PAMI,
  volume = {47},
  number = {5},
  pages = {3424--3439},
  year = {2025},
  doi = {10.1109/TPAMI.2025.3532688}
}

@article{otter_tpami2025,
  title = {{Otter}: A Multi-Modal Model with In-Context Instruction Tuning},
  author = {Li, Bo and Zhang, Yuanhan and Chen, Liangyu and Wang, Jinghao and Pu, Fanyi and Cahyono, Joshua Adrian and Yang, Jingkang and Li, Chunyuan and Liu, Ziwei},
  journal = PAMI,
  volume = {47},
  number = {9},
  pages = {7543--7557},
  year = {2025},
  doi = {10.1109/TPAMI.2025.3571946}
}

@article{gen3dvlm_tpami2025,
  title = {General {3D} Vision-Language Model with Fast Rendering and Pre-Training Vision-Language Alignment},
  author = {Liu, Kangcheng and Liu, Yong-Jin and Chen, Baoquan},
  journal = PAMI,
  volume = {47},
  number = {9},
  pages = {7352--7368},
  year = {2025},
  doi = {10.1109/TPAMI.2025.3566593}
}

@article{contrastive_videoqa_tpami2023,
  title = {Contrastive Video Question Answering via Video Graph Transformer},
  author = {Xiao, Junbin and Zhou, Pan and Yao, Angela and Li, Yicong and Hong, Richang and Yan, Shuicheng and Chua, Tat-Seng},
  journal = PAMI,
  volume = {45},
  number = {11},
  pages = {13265--13280},
  year = {2023},
  doi = {10.1109/TPAMI.2023.3292266}
}

@article{invariant_grounding_videoqa_tpami2025,
  title = {Transformer-Empowered Invariant Grounding for Video Question Answering},
  author = {Li, Yicong and Wang, Xiang and Xiao, Junbin and Ji, Wei and Chua, Tat-Seng},
  journal = PAMI,
  volume = {47},
  number = {11},
  pages = {9510--9522},
  year = {2025},
  doi = {10.1109/TPAMI.2023.3303451}
}

@article{jm3d_tpami2025,
  title = {{JM3D} and {JM3D-LLM}: Elevating {3D} Representation with Joint Multi-Modal Cues},
  author = {Ji, Jiayi and Wang, Haowei and Wu, Changli and Ma, Yiwei and Sun, Xiaoshuai and Ji, Rongrong},
  journal = PAMI,
  volume = {47},
  number = {4},
  pages = {2475--2492},
  year = {2025},
  doi = {10.1109/TPAMI.2024.3523675}
}

@article{crossmodal_causal_vqa_tpami2023,
  title = {Cross-Modal Causal Relational Reasoning for Event-Level Visual Question Answering},
  author = {Liu, Yang and Li, Guanbin and Lin, Liang},
  journal = PAMI,
  volume = {45},
  number = {10},
  pages = {11624--11641},
  year = {2023},
  doi = {10.1109/TPAMI.2023.3284038}
}

\clearpage

\end{document}